\documentclass[10pt]{article} 
\usepackage[preprint]{tmlr}

\usepackage{amsmath,amsfonts,bm}

\def\eqref#1{equation~\ref{#1}}

\def\1{\bm{1}}

\DeclareMathAlphabet{\mathsfit}{\encodingdefault}{\sfdefault}{m}{sl}
\SetMathAlphabet{\mathsfit}{bold}{\encodingdefault}{\sfdefault}{bx}{n}

\usepackage{hyperref}
\usepackage{url}
\usepackage{graphicx}
\usepackage{subcaption}
\usepackage{multirow}
\usepackage{wrapfig}
\usepackage{booktabs}

\title{Resolving Mixed Single-Photon LiDAR Returns for Foreground-View and Hidden Scene Reconstruction}

\author{\name Ziting Wen \email wenzt@sustech.edu.cn \\
      \addr School of Automation and Intelligent Manufacturing, Southern University of Science and Technology
      \AND
      \name Runrong Deng \email 12532835@mail.sustech.edu.cn \\
      \addr School of Automation and Intelligent Manufacturing, Southern University of Science and Technology
      \AND
      \name Zili Zhang \email zhangzili1995@shu.edu.cn \\
      \addr School of Mechatronic Engineering and Automation, Shanghai University
      \AND
      \name Haitao Zheng \email zhenghaitao@shu.edu.cn \\
      \addr School of Mechatronic Engineering and Automation, Shanghai University
      \AND
      \name Yuecong Xu \email yc.xu@nus.edu.sg \\
      \addr Department of Electrical and Computer Engineering, National University of Singapore
      \AND
      \name Xiaoqiang Ren \email xqren@shu.edu.cn \\
      \addr School of Mechatronic Engineering and Automation, Shanghai University
      \AND
      \name Guodong Shi \email guodong.shi@sydney.edu.au \\
      \addr Australian Centre for Robotics, University of Sydney
      \AND
      \name Kemi Ding \email dingkm@sustech.edu.cn \\
      \addr School of Automation and Intelligent Manufacturing, Southern University of Science and Technology}

\begin{document}

\maketitle

\begin{abstract}
Partially transmissive screens and protective covers are common in robotic inspection, but they create mixed LiDAR returns from both the foreground material and the scene behind it. Conventional peak-based LiDAR usually discards weak hidden returns, while single-photon LiDAR records time-resolved histograms that preserve attenuated and overlapping echoes. However, existing transient reconstruction methods typically fit a single scene representation to the measured waveform. Under occlusion, weak or nearby foreground--hidden echoes can form a broad peak or subtle shoulder. Because such waveforms can also be explained by a displaced single surface or a thick density distribution, accurate transient fitting does not necessarily imply correct geometry. We propose a state-aware framework for foreground-view and hidden scene reconstruction from occluded single-photon histograms. For each ray, we estimate local echo evidence, identifying no reliable surface evidence, single-return evidence, or two returns. The inferred echo state routes supervision for a two-head neural field: all rays constrain waveform reconstruction, while reliable anchors provide geometry localization. We also introduce a real paired single-photon LiDAR occlusion dataset with occluded and clean captures at fixed poses. Experiments on a real dataset show improved hidden scene depth and point-cloud accuracy over baselines. Our results demonstrate single-photon layered reconstruction as a practical route for 3D perception through partially transmissive occluders.
\end{abstract}

\section{Introduction}

\begin{figure}[!htbp]
    \centering
    \includegraphics[width=\columnwidth]{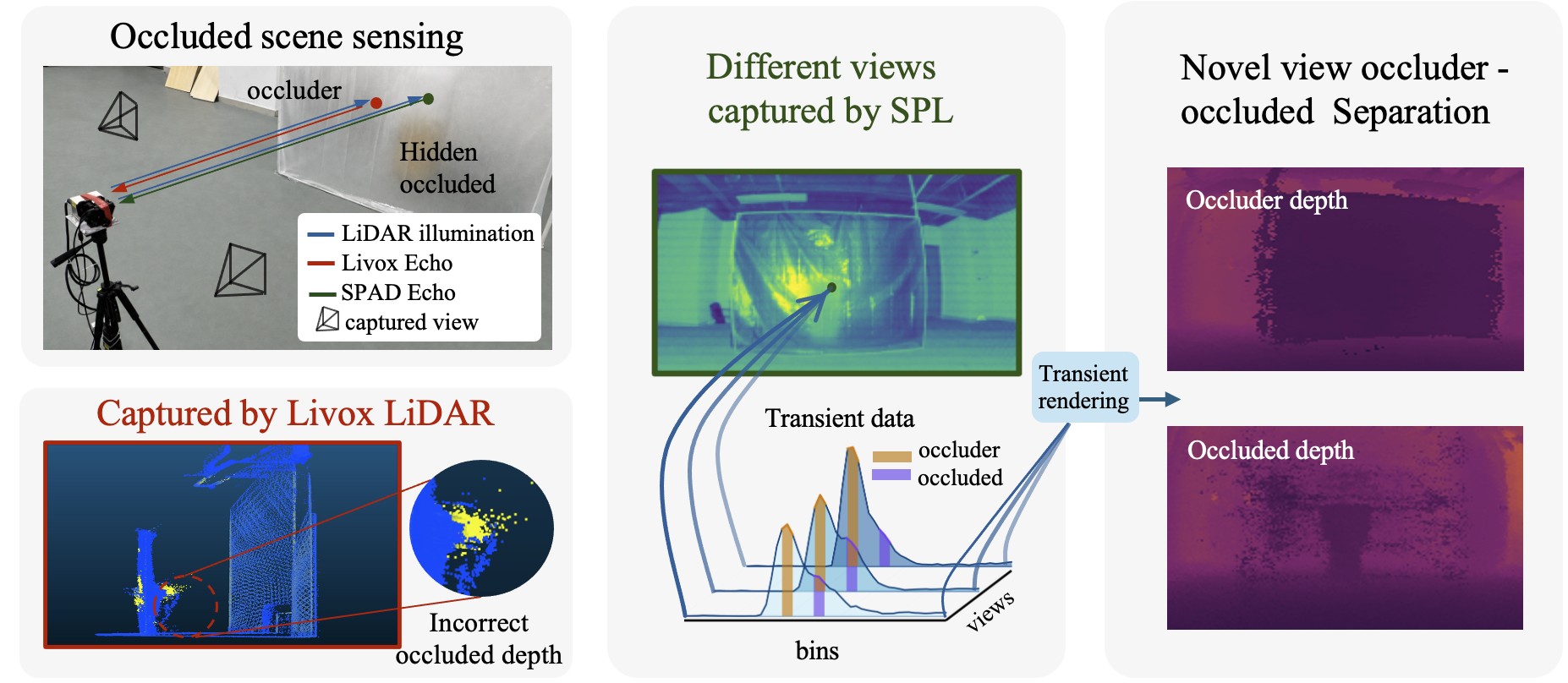}
    \caption{Task overview. An occluder produces mixed foreground and hidden returns along the same ray. Conventional peak-based LiDAR often stops at the occluder, while SPAD transient histograms preserve weak delayed hidden echoes. We use multi-view occluded SPAD measurements to reconstruct separate foreground-view and hidden scene geometry. }
    \label{fig:intro}
\end{figure}

Recovering 3D geometry through partially transmissive occluders is important for inspection, monitoring, and robotic perception in screened or protected environments~\citep{almadhoun2016survey,border2024surface}. Screens, curtains, and protective covers are often neither fully opaque nor optically transparent: they reflect part of the emitted light while transmitting a weaker component to the scene behind them. As a result, a single sensor ray may contain direct returns from both the foreground
occluder and the hidden surface. This turns occluded sensing into a
layered reconstruction problem: the goal is to recover a
foreground-view reconstruction containing the occluder and directly
visible background, together with a hidden scene reconstruction
containing the surfaces behind the occluder and the same visible
background. Conventional peak-based LiDAR pipelines are limited in this setting. By reducing each measurement to one or a few extracted ranges, they may terminate at the strong foreground return, merge nearby foreground and hidden returns, or discard the attenuated hidden return during range extraction. Single-photon LiDAR (SPL) provides a more suitable measurement. SPAD sensitivity helps detect small numbers of returning photons from hidden surfaces, while time-correlated single-photon counting records their arrival times as a full transient histogram~\citep{kirmani2014first,rapp2020advances,scheuble2025lidar}. Instead of committing to a single depth, the histogram preserves weak secondary echoes, asymmetric peaks, and shoulder-like structures produced by multiple surfaces along the same line of sight, as shown in
Fig.~\ref{fig:intro}. These properties make SPL a
promising modality for layered 3D reconstruction through partially
transmissive occlusion.

Recent transient reconstruction methods show that raw single-photon
histograms can supervise neural 3D reconstruction and novel-view
transient synthesis~\citep{malik2023transient,
luo2025transientangelo,scheuble2025transient}. These methods typically
optimize a single scene representation by matching the rendered
transient to the measured waveform. This is effective when each ray is
dominated by one surface, but becomes under-constrained under layered
occlusion. A foreground occluder and a hidden surface can generate
nearby echoes along the same ray; when the hidden return is weak or
temporally close to the foreground return, the histogram may appear as
a broad peak, an asymmetric tail, or a subtle shoulder. Such a waveform
can be explained by the true surface pair, but also by a displaced
single surface or a thick density distribution. Thus, accurate
transient fitting does not necessarily imply correct
foreground--hidden geometry. The geometric evidence contained in the histogram is also
ray-dependent. Some rays provide no reliable surface evidence, some
support only a single return, and some support an ordered
foreground--hidden return pair. Direct peak extraction can miss weak
hidden returns or assign them to the wrong layer, while uniformly
forcing all rays to explain two layers can introduce unsupported
geometry. The key challenge is therefore to use each transient
histogram only to the extent supported by its local echo evidence, and
to integrate these heterogeneous cues into a globally consistent
foreground--hidden reconstruction.

We address this challenge with a state-aware layered reconstruction
framework. For each ray, we estimate local echo evidence, identifying whether the measured
histogram provides no reliable anchor evidence, a single-return anchor,
or ordered foreground--hidden anchors. The inferred echo state routes
the supervision for a neural field with foreground and
hidden heads: all rays contribute to waveform fitting, while reliable
anchors activate geometry localization. This evidence-dependent supervision reduces the layer-assignment ambiguity of transient fitting and allows the two heads to be rendered independently as foreground-view and
hidden scene reconstructions. To evaluate this setting, we capture a real paired multi-view
single-photon LiDAR occlusion dataset. For each scene and fixed sensor
pose, we acquire an occluded measurement and a clean reference captured
after removing the occluder while keeping the hidden object and
background unchanged. Reconstruction uses only the occluded
measurements, while the paired clean captures are reserved for
evaluation. The dataset contains 30 scenes with three partially
transmissive materials and 10 hidden-object categories, providing a
real-data benchmark for foreground-view and hidden scene
reconstruction.

Our contributions are: (1) To the best of our knowledge, we present the first study of foreground-view and hidden scene reconstruction from mixed SPL histograms using only occluded measurements, and show that accurate transient fitting alone is insufficient to resolve layered geometry. (2) We propose a state-aware two-head neural field that routes per-ray echo evidence into waveform, geometry-localization, and foreground–hidden partition supervision. (3) We introduce and will release the first real paired multi-view SPL dataset for occlusion reconstruction, including occluded/clean captures at fixed poses and hidden-object masks.

\section{Related Work}

\textbf{Occlusion-aware 3D reconstruction.}
Existing occlusion-aware 3D reconstruction methods mainly rely on two sources of information: visible evidence from other viewpoints and priors for completing unobserved regions. NeRF-based methods follow this principle by modeling visibility, down-weighting unreliable observations, removing foreground distractors, or aggregating unoccluded background evidence across views
\citep{martin2021nerf,sabour2023robustnerf,
ren2024nerf,zhu2023occlusion}. Recent 3DGS-based methods extend
the same idea with explicit Gaussian primitives, using generative priors or co-visibility partitioning to improve reconstruction and rendering in partially observed scenes~\citep{sun2024occfusion,liu2025declutternerf}. Although these works differ in representation and optimization, their measurements remain visibility-limited: hidden geometry must either be seen from some viewpoints or inferred from priors. In our setting, a partially transmissive foreground and a hidden surface can both contribute to the same sensor ray. The hidden scene is therefore not only missing from the image; it may be encoded as weak delayed photons in the single-photon LiDAR histogram. This motivates our use of time-resolved histograms for separating foreground and hidden geometry.

\textbf{LiDAR and single-photon transient reconstruction.}
Neural LiDAR methods extend neural scene representations to active
sensing by modeling range, beam effects, or secondary returns~\citep{zhang2024nerf,wu2024dynamic,
jianggs}. These methods mainly operate on
extracted point clouds. Single-photon LiDAR instead
records a photon-count histogram for each ray, preserving weak and
overlapping echoes~\citep{scheuble2025lidar}. Classical single-photon imaging methods model photon statistics for depth, reflectivity, and multi-return recovery under obscurants or semi-transparent surfaces~\citep{kirmani2014first,rapp2017few,halimi2017restoration,plosz2023real,maccarone2023submerged}. Recent neural transient methods further optimize scene representations from
raw transient measurements~\citep{malik2023transient,luo2025transientangelo,
scheuble2025transient,malik2024flying}. While these works show the value of histogram supervision, they typically fit a single scene representation to the total measured transient. Under occlusion, nearby or attenuated foreground and hidden returns can be explained by a thickened or wrongly assigned surface. We therefore combine state-aware echo evidence and layer supervision to recover foreground-view and hidden scene geometry from mixed histograms.

\section{Method}
\label{sec:method}

\begin{figure*}[!htbp]
    \centering
    \includegraphics[width=0.97\textwidth]{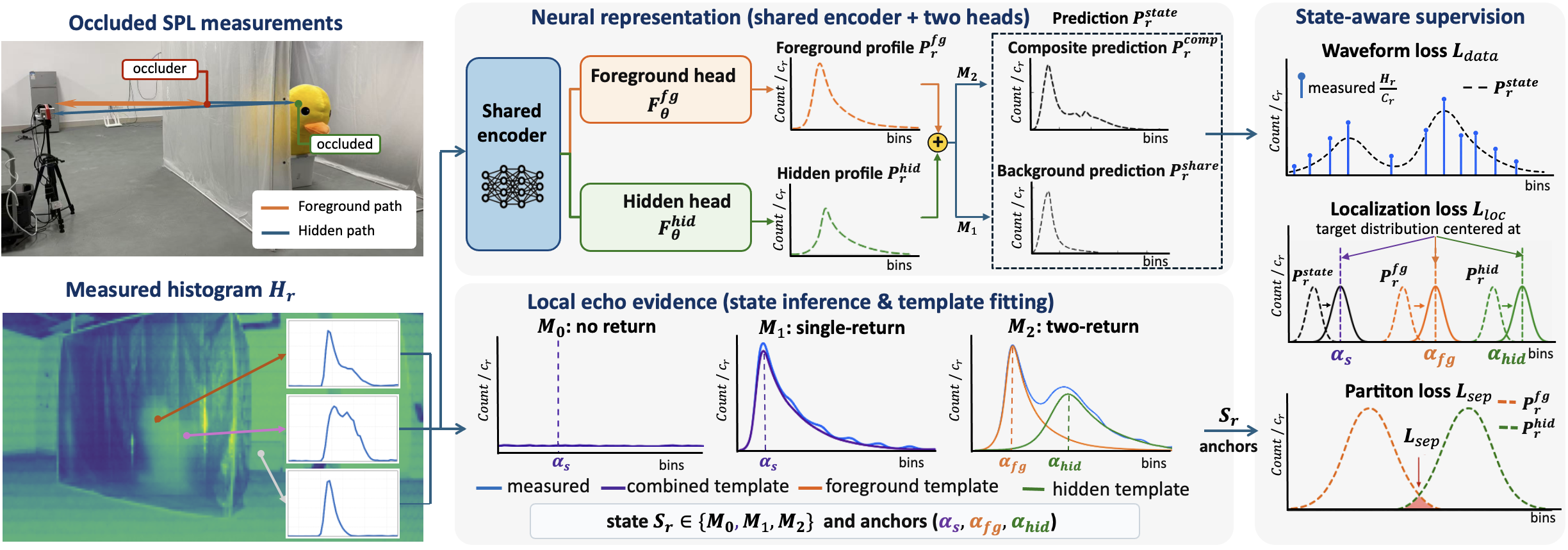}
    \caption{Overview of our state-aware single-photon layered reconstruction framework. From occluded histograms, we infer per-ray echo states and temporal anchors via template fitting, then route waveform, localization, and partition supervision for a two-head neural field. The two heads are rendered independently to recover foreground-view and hidden scene geometry.}
    \label{fig:framew}
\end{figure*}

\subsection{Overview}

We consider a static scene observed by a SPL through a
partially transmissive foreground occluder. Each ray records a
time-resolved photon-count histogram that may contain mixed returns
from the occluder and the scene behind it. Given only the occluded
histograms and the LiDAR poses, our goal is to recover two
geometry outputs: a foreground-view reconstruction, containing the
occluder and the directly visible background, and a hidden scene
reconstruction, containing surfaces behind the occluder and the same
visible background. The histogram is not a direct layer label: foreground and hidden returns are mixed by the system response and the single-photon detection process, so nearby echoes can overlap and strong early returns can suppress later ones. We therefore treat each ray according to the evidence it supports: ordered foreground--hidden returns, a single reliable return, or no reliable surface anchor.

Our method adopts the local-to-global reconstruction strategy illustrated in Fig.~\ref{fig:framew}. We first analyze each histogram under the same calibrated response templates, obtaining an echo state and any supported temporal anchors. A shared neural field with separate foreground and hidden heads then integrates these local cues across views. The echo state determines how each ray supervises the two heads: rays without reliable anchors only fit the measured waveform, single-return rays preserve shared background geometry, and two-return rays separately constrain the occluder and hidden scene. A factorized objective then turns the local anchors into constraints on geometry localization and head assignment.

\subsection{Single-Photon Transient Formation}
\label{subsec:single_photon_formation}

A SPL combines pulsed illumination, SPAD detection, and
time-correlated photon counting. For each sensing ray \(r\), the laser
repeatedly emits a pulse with temporal profile \(s(t)\), and the
detector records photon arrival times relative to the emitted pulses.
Let \(\widetilde{\Phi}_r(t)\) denote the latent photon-arrival flux at
the detector input that gives rise to the measured histogram. If the
ray observes a single surface at range \(z_r\) with return strength
\(\alpha_r\), this arrival flux can be written as a delayed and scaled
pulse~\citep{shin2015photon},
\begin{equation}
\widetilde{\Phi}_r(t)
=
\eta \alpha_r
s\!\left(t-\frac{2z_r}{c}\right)
+
\widetilde{\Phi}_r^{\mathrm{amb}}(t),
\label{eq:single_surface_flux}
\end{equation}
where \(\eta\) is the detection efficiency, \(c\) is the speed of
light, and \(\widetilde{\Phi}_r^{\mathrm{amb}}\) accounts for ambient
photons and dark counts. The pulse delay encodes range, while its
magnitude depends on the returned signal strength. In practice, the time axis is discretized into \(B\) bins. After many
illumination cycles, the recorded photon arrivals form a histogram
\(\mathbf H_r=\{H_{r,b}\}_{b=1}^{B}\), where \(H_{r,b}\) is the
number of detections in bin \(b\), and
\(C_r=\sum_b H_{r,b}\) is the total recorded count. Throughout the
paper, a subscript \(b\) denotes the value of a binned temporal profile
at bin \(b\). Unlike a peak-extracted range measurement, the full
histogram preserves weak and overlapping returns along the same line
of sight, which is essential for rays that contain both a foreground
occluder return and a hidden scene return.

After discretization, the latent arrival flux becomes
\(\widetilde{\boldsymbol{\Phi}}_r
=\{\widetilde{\Phi}_{r,b}\}_{b=1}^{B}\), where
\(\widetilde{\Phi}_{r,b}\) is the expected number of photon arrivals
in bin \(b\) before first-photon selection and dead-time competition.
Under Poisson arrival statistics, the probability of at least one
arrival in bin \(b\) is \(1-e^{-\widetilde{\Phi}_{r,b}}\). After a
photon is detected, the SPAD is temporarily inactive, so an early
strong return can suppress detections in later bins within the
dead-time window. Following previous works
~\citep{coates1968correction,kirmani2014first}, the probability that a
recorded detection falls in bin \(b\) is
\begin{equation}
\widetilde{P}_{r,b}
\propto
\left(1-e^{-\widetilde{\Phi}_{r,b}}\right)
e^{-\sum_{j\in\mathcal D_b}\widetilde{\Phi}_{r,j}} .
\label{eq:windowed_first_photon}
\end{equation}
where \(\mathcal D_b\) contains the preceding bins covered by the
dead-time window. The probabilities are normalized over the modeled
histogram range so that \(\sum_{b=1}^{B}\widetilde{P}_{r,b}=1\).
We denote this detector mapping by
\(\widetilde{\mathbf P}_r = \mathcal S_{\mathrm{win}}(\widetilde{\boldsymbol{\Phi}}_r)\). Here, \(\widetilde{\mathbf P}_r\) is the arrival-time distribution of a recorded detection after dead-time competition. Since a histogram accumulates \(C_r\) such detections over repeated illumination cycles, we model
the bin counts conditionally as
\begin{equation}
\mathbf H_r \mid C_r
\sim
\operatorname{Multinomial}(C_r,\widetilde{\mathbf P}_r).
\label{eq:histogram_likelihood}
\end{equation}

We next specify the neural-field prediction \(\boldsymbol{\Phi}_r\) of the latent photon-arrival flux \(\widetilde{\boldsymbol{\Phi}}\). Following time-resolved LiDAR
rendering~\citep{malik2023transient}, we sample points along ray \(r\) as
\(\mathbf x_i=\mathbf o_r+c t_i\mathbf d_r\), where
\(\mathbf o_r\) is the ray origin, \(\mathbf d_r\) is the unit ray
direction, and \(t_i\) is the one-way propagation time of the sample.
The scene is represented by a shared spatial encoder with two heads, \(F_\theta^{\mathrm{fg}}\) and \(F_\theta^{\mathrm{hid}}\), corresponding to the foreground-view and hidden scene outputs, respectively. In occluded regions, they represent the occluder and behind-occluder surfaces, while directly visible surfaces are retained in both. For each sample, head \(k\in\{\mathrm{fg},\mathrm{hid}\}\) predicts a density and return amplitude,
\((\sigma_i^k,\rho_i^k)=F_\theta^k(\mathbf x_i)\). With samples ordered by increasing one-way time \(t_i\), the accumulated transmittance of head \(k\) before sample \(i\) is \(\mathcal T_i^k=\exp(-\sum_{j<i}\sigma_j^k\delta_j)\). We define the
NeRF-style ray-termination weight as \(w_i^k=\mathcal T_i^k(1-\exp(-\sigma_i^k\delta_i))\). The
pre-response transient of head \(k\) is then \(\tau_{r,b}^{k}=\sum_{i:\,\beta(2t_i)=b}w_i^k\rho_i^k\), where \(\beta(2t_i)\) maps the round-trip time of sample \(i\) to its histogram bin. Thus, \(w_i^k\) describes ray-termination geometry, while \(\boldsymbol{\tau}_r^k\) is the time-resolved transient before temporal broadening by the LiDAR system.  The foreground and hidden pre-response transients are temporally broadened by calibrated effective temporal responses and then summed
with the ambient-and-dark component:
\begin{equation}
\boldsymbol{\Phi}_r
=
\boldsymbol{\kappa}^{\mathrm{fg}} *
\boldsymbol{\tau}_r^{\mathrm{fg}}
+
\boldsymbol{\kappa}^{\mathrm{hid}} *
\boldsymbol{\tau}_r^{\mathrm{hid}}
+
\boldsymbol{\Phi}_r^{\mathrm{amb}} .
\label{eq:layered_flux}
\end{equation}
Here, \(*\) denotes discrete convolution along the temporal-bin axis.
The responses \(\boldsymbol{\kappa}^{\mathrm{fg}}\) and
\(\boldsymbol{\kappa}^{\mathrm{hid}}\) model the effective temporal
broadening of foreground and hidden returns, respectively.
The term \(\boldsymbol{\Phi}_r^{\mathrm{amb}}\) denotes ambient photons
and dark counts, modeled as a uniform temporal component with a fitted
amplitude. Applying the windowed first-photon mapping gives the
composite detection-time distribution
\(\mathbf P_r^{\mathrm{comp}}
=
\mathcal S_{\mathrm{win}}(\boldsymbol{\Phi}_r)\).

\subsection{Local Evidence and State-Aware Reconstruction}
\label{subsec:state_aware_supervision}

Local echo evidence summarizes the surface structure supported by a
single occluded histogram before multi-view reconstruction. For each
ray, it provides an echo state, temporal anchors, and reliability
scores, indicating whether the histogram supports no surface anchor,
one temporal anchor, or an ordered foreground--hidden anchor pair.

We estimate this evidence by fitting calibrated single-return
templates under the windowed first-photon mapping. We use two effective temporal responses: a response \(\boldsymbol{\kappa}^{\mathrm{fg}}\) for
unobstructed or foreground-material returns, and a hidden response
\(\boldsymbol{\kappa}^{\mathrm{hid}}\) for later hidden-side returns.
Let \(\mathbf R^\ell(a)\) denote response
\(\ell\in\{\mathrm{fg},\mathrm{hid}\}\) shifted to temporal bin \(a\), and let \(\mathbf u\) denote a temporally uniform ambient-and-dark-count template. The variables \(a_s,a_f,a_h\) are fitted temporal centers, and the \(\alpha\)'s are nonnegative fitted amplitudes. For each ray, we compare three expected-arrival explanations:
\begin{equation}
\begin{aligned}
\boldsymbol{\Phi}_r^{\mathcal M_0}
&= \alpha_{amb}\mathbf u,\\
\boldsymbol{\Phi}_r^{\mathcal M_1}
&= \alpha_s\mathbf R^{\mathrm{fg}}(a_s)+\alpha_{amb}\mathbf u,\\
\boldsymbol{\Phi}_r^{\mathcal M_2}
&= \alpha_f\mathbf R^{\mathrm{fg}}(a_f)
 + \alpha_h\mathbf R^{\mathrm{hid}}(a_h)
 + \alpha_{amb}\mathbf u.
\end{aligned}
\label{eq:template_models}
\end{equation}

These explanations correspond to no-return, single-return, and two-return histograms. For each explanation
\(\mathcal M_m\), we optimize its amplitudes and temporal centers,
convert the fitted expected-arrival profile to a detection-time
distribution
\(\mathbf P_r^m
=
\mathcal S_{\mathrm{win}}(\boldsymbol{\Phi}_r^{\mathcal M_m})\),
and fit the measured histogram using the conditional multinomial
negative log-likelihood. A BIC-style penalized likelihood~\citep{schwarz1978estimating} then selects
the echo state \(s_r\in\{0,1,2\}\), favoring the simplest explanation
that fits the observed waveform. 

The selected model determines both the local cues and the supervision
route used for reconstruction. If \(\mathcal M_0\) is selected, the
ray provides no reliable temporal anchor. If \(\mathcal M_1\) is
selected, it provides a single-surface anchor \(a_r^s=a_s\) with
confidence \(c_r^s\). If \(\mathcal M_2\) is selected, it provides
ordered anchors \(a_r^{\mathrm{fg}}=a_f\) and
\(a_r^{\mathrm{hid}}=a_h\), together with hidden-anchor reliability
\(\gamma_r^{\mathrm{hid}}\). Rays assigned to \(\mathcal M_0\) and \(\mathcal M_2\) use the
composite prediction \(\mathbf P_r^{\mathrm{comp}}\). The former
receive no anchor supervision, whereas the ordered anchors of the
latter supervise the foreground and hidden heads separately. Under
our output definition, the foreground-view and hidden scene
reconstructions share the unoccluded scene background. We therefore
use reliable single-return rays to preserve this shared component in
both heads and apply the same anchor \(a_r^s\) to each head. Each head
is rendered independently using the visible single-return response,
\(\mathbf P_r^{k}=\mathcal S_{\mathrm{win}}
(\boldsymbol{\kappa}^{\mathrm{fg}}*\boldsymbol{\tau}_r^{k}
+\boldsymbol{\Phi}_r^{\mathrm{amb}})\), where
\(k\in\{\mathrm{fg},\mathrm{hid}\}\), and the corresponding
shared-background prediction is
\(\mathbf P_r^{\mathrm{share}}
=\tfrac{1}{2}(\mathbf P_r^{\mathrm{fg}}
+\mathbf P_r^{\mathrm{hid}})\).
For a lower-confidence single-return fit, we instead use
\(\mathbf P_r^{\mathrm{comp}}\) and impose the anchor only on the
combined geometry of the two heads.

The waveform prediction is routed according to the inferred echo
state: reliable single-return rays use
\(\mathbf P_r^{\mathrm{share}}\), whereas no-anchor rays,
two-return rays, and uncertain single-return rays use
\(\mathbf P_r^{\mathrm{comp}}\). We denote the selected prediction by
\(\mathbf P_r^{\mathrm{state}}\). Following the conditional histogram
likelihood in Eq.~\eqref{eq:histogram_likelihood}, the waveform
reconstruction loss is
\begin{equation}
\mathcal L_{\mathrm{data}}
=
-\sum_r\sum_{b=1}^{B}
\frac{H_{r,b}}{C_r}
\log
\left(
P_{r,b}^{\mathrm{state}}+\epsilon
\right).
\label{eq:state_aware_data_loss}
\end{equation}
Here, \(\epsilon\) is a small constant for numerical
stability. Thus, every ray contributes to waveform fitting, while the inferred
echo state determines whether it provides shared-background supervision or separate foreground--hidden
supervision.

\subsection{Layer Supervision and Geometry Recovery}
\label{subsec:factorized_supervision}

The echo state determines which anchors are available for each ray.
We convert these anchors into two geometric losses: a localization
loss that places geometry near the inferred temporal anchors, and a
partition loss that assigns two-return geometry to the foreground or
hidden head. For geometric supervision, we bin the NeRF weights along the histogram
axis as
\(W_{r,b}^{k}=\sum_{i:\,\beta(2t_i)=b} w_i^k\),
where \(k\in\{\mathrm{fg},\mathrm{hid}\}\). The resulting profile
\(\mathbf W_r^k\) describes where head \(k\) places ray-termination
geometry along ray \(r\).

\textbf{Geometry anchoring.}
For a temporal anchor \(a\), let \(\mathbf q_a\) be a narrow target
distribution centered at \(a\). We define
\begin{equation}
\mathcal A(\mathbf W,a)
=
\mathcal L_{\mathrm{pres}}(\mathbf W,a)
+
\mathcal W_1
\left(
\operatorname{norm}(\mathbf W),
\mathbf q_a
\right),
\label{eq:anchor_objective}
\end{equation}
where \(\mathcal W_1\) is the one-dimensional Wasserstein distance.
The presence term encourages sufficient local geometry around the
anchor; in our implementation it penalizes
\(\max(0,\mu-M_a(\mathbf W))^2\), where
\(M_a(\mathbf W)=\sum_b \mathbf q_a(b)W_b\) is the geometry mass near \(a\). For two-return rays, the localization loss is
\begin{equation}
\mathcal L_{\mathrm{loc}}(r)
=
\mathcal A(\mathbf W_r^{\mathrm{fg}},a_r^{\mathrm{fg}})
+
\gamma_r^{\mathrm{hid}}
\mathcal A(\mathbf W_r^{\mathrm{hid}},a_r^{\mathrm{hid}}).
\label{eq:two_return_localization}
\end{equation}
For reliable single-return rays, the same anchor \(a_r^s\) supervises
both heads. For uncertain single-return rays, the anchor supervises
only the summed geometry
\(\mathbf W_r^{\mathrm{fg}}+\mathbf W_r^{\mathrm{hid}}\). 

\textbf{Head partition.}
For a two-return ray, the anchors define a temporal split
\(m_r=(a_r^{\mathrm{fg}}+a_r^{\mathrm{hid}})/2\). We use a
hidden-side gate
\(g_r(b)=sigmoid((b-m_r)/\tau_g)\), where \(g_r(b)\) is close to zero
on the foreground side and close to one on the hidden side. With
\(\bar{\mathbf W}_r^k=\operatorname{norm}(\mathbf W_r^k)\), the
responsibility loss is
\begin{equation}
\mathcal L_{\mathrm{resp}}(r)
=
\sum_b
\left[
\bar W_{r,b}^{\mathrm{fg}} g_r(b)
+
\bar W_{r,b}^{\mathrm{hid}}(1-g_r(b))
\right].
\label{eq:responsibility_loss}
\end{equation}
We also penalize overlapping geometry:
\begin{equation}
\mathcal L_{\mathrm{ov}}(r)
=
\sum_b
\frac{
4W_{r,b}^{\mathrm{fg}}W_{r,b}^{\mathrm{hid}}
}{
\left(
W_{r,b}^{\mathrm{fg}}+
W_{r,b}^{\mathrm{hid}}+
\epsilon
\right)^2
}.
\label{eq:overlap_loss}
\end{equation}
The partition loss is
\begin{equation}
\mathcal L_{\mathrm{sep}}(r)
=
\mathbb{I}_{\{s_r=2\}}\,
\gamma_r^{\mathrm{hid}}
\left(
\mathcal L_{\mathrm{resp}}(r)
+
\mathcal L_{\mathrm{ov}}(r)
\right).
\label{eq:partition_objective}
\end{equation}
Here, \(\mathcal L_{\mathrm{resp}}\) assigns the earlier and later
anchored regions to the foreground and hidden heads, respectively,
while \(\mathcal L_{\mathrm{ov}}\) discourages both heads from
explaining the same temporal bin. The final training objective is
\begin{equation}
\mathcal L
=
\mathcal L_{\mathrm{data}}
+
\lambda_{\mathrm{loc}}
\sum_r \mathcal L_{\mathrm{loc}}(r)
+
\lambda_{\mathrm{sep}}
\sum_r \mathcal L_{\mathrm{sep}}(r).
\label{eq:total_objective}
\end{equation}
Each ray contributes through the waveform loss in
Eq.~\eqref{eq:state_aware_data_loss}, while localization and partition
losses are activated according to the inferred echo state. During
training, we sample rays across no return, single-return, and
two-return states so that the sparse two-return supervision remains
effective. At inference, the two heads are rendered independently. The depth of
head \(k\) is taken from the peak of its geometry profile,
\(D_r^k=\operatorname{range}(\arg\max_b W_{r,b}^k)\), and the
foreground and hidden depth maps are projected into separate
point clouds.

\section{Experiments}

\begin{table*}[!htbp]
\centering
\footnotesize
\setlength{\tabcolsep}{1.8pt}
\renewcommand{\arraystretch}{0.90}
\caption{Quantitative comparison on the captured multi-view occlusion LiDAR dataset. CD and L1 are lower better, while F1@2 is higher better. Best results within each occluder group are in bold, and second-best results are underlined.}
\label{tab:main_results}
\begin{tabular}{@{}l@{\hspace{0.5em}}l@{\hspace{0.6em}}ccc@{\hspace{0.7em}}ccc@{\hspace{0.7em}}ccc@{}}
\toprule
\multirow{2}{*}{Occluder} & \multirow{2}{*}{Method}
& \multicolumn{3}{c}{Foreground}
& \multicolumn{3}{c}{Hidden}
& \multicolumn{3}{c}{Mask-on-Hidden} \\
\cmidrule(lr){3-5} \cmidrule(lr){6-8} \cmidrule(l){9-11}
& & CD (m) $\downarrow$ & F1@2 $\uparrow$ & L1 (m) $\downarrow$
& CD (m) $\downarrow$ & F1@2 $\uparrow$ & L1 (m) $\downarrow$
& CD (m) $\downarrow$ & F1@2 $\uparrow$ & L1 (m) $\downarrow$ \\
\midrule
\multirow[c]{6}{*}{Black mesh}
& DyNFL & 0.373 & 0.692 & 1.656 & 0.408 & 0.642 & 1.904 & 0.375 & 0.499 & 1.197 \\
& Transientangelo & 0.194 & 0.841 & 1.768 & 0.155 & \underline{0.857} & 1.244 & 0.495 & 0.525 & 0.840 \\
& TransientNeRF & 0.299 & 0.748 & 0.986 & 0.179 & 0.838 & 1.092 & 0.415 & \underline{0.546} & 0.967 \\
& Flying with Photons & 0.252 & 0.634 & 1.064 & 0.241 & 0.705 & 1.056 & 0.578 & 0.453 & 1.386 \\
& Multilayer Imaging & \textbf{0.076} & \textbf{0.952} & \textbf{0.427} & \underline{0.134} & 0.846 & \underline{0.873} & \underline{0.245} & 0.491 & \underline{0.642} \\
& Ours & \underline{0.134} & \underline{0.848} & \underline{0.490} & \textbf{0.107} & \textbf{0.907} & \textbf{0.496} & \textbf{0.118} & \textbf{0.886} & \textbf{0.342} \\
\midrule
\multirow[c]{6}{*}{Mosquito net}
& DyNFL & 0.137 & 0.879 & 0.734 & 0.191 & 0.811 & 1.105 & 0.240 & 0.584 & 0.794 \\
& Transientangelo & 0.199 & 0.846 & 1.883 & \underline{0.160} & \underline{0.861} & 1.261 & 0.550 & 0.399 & 0.956 \\
& TransientNeRF & 0.309 & 0.724 & 1.100 & 0.186 & 0.836 & \underline{1.004} & 0.512 & 0.434 & 0.987 \\
& Flying with Photons & 0.276 & 0.688 & 1.417 & 0.304 & 0.691 & 1.281 & 0.665 & 0.413 & 0.957 \\
& Multilayer Imaging & \underline{0.078} & \underline{0.951} & \underline{0.428} & 0.170 & 0.805 & 0.995 & \underline{0.243} & \underline{0.709} & \underline{0.617} \\
& Ours & \textbf{0.076} & \textbf{0.954} & \textbf{0.361} & \textbf{0.099} & \textbf{0.926} & \textbf{0.425} & \textbf{0.139} & \textbf{0.867} & \textbf{0.414} \\
\midrule
\multirow[c]{6}{*}{PVC curtain}
& DyNFL & \underline{0.135} & \underline{0.877} & 0.820 & 0.184 & 0.813 & 1.096 & \underline{0.211} & 0.629 & \underline{0.653} \\
& Transientangelo & 0.292 & 0.765 & 2.126 & 0.171 & \underline{0.859} & 1.325 & 0.530 & \underline{0.781} & 0.919 \\
& TransientNeRF & 0.407 & 0.693 & 1.368 & 0.192 & 0.854 & 1.095 & 0.573 & 0.685 & 0.834 \\
& Flying with Photons & 0.202 & 0.765 & 1.201 & 0.241 & 0.748 & 1.216 & 0.555 & 0.531 & 1.466 \\
& Multilayer Imaging & \textbf{0.095} & \textbf{0.920} & \textbf{0.601} & \underline{0.157} & 0.815 & \underline{0.999} & 0.321 & 0.396 & 0.750 \\
& Ours & 0.144 & 0.848 & \underline{0.639} & \textbf{0.119} & \textbf{0.900} & \textbf{0.499} & \textbf{0.176} & \textbf{0.800} & \textbf{0.569} \\
\midrule
\multirow[c]{2}{*}{Avg.}
& DyNFL & 0.215 & 0.816 & 1.070 & 0.261 & 0.755 & 1.368 & 0.275 & \underline{0.571} & 0.881 \\
& Transientangelo & 0.229 & 0.817 & 1.926 & 0.162 & \underline{0.859} & 1.277 & 0.525 & 0.568 & 0.905 \\
& TransientNeRF & 0.338 & 0.721 & 1.151 & 0.186 & 0.843 & 1.064 & 0.500 & 0.555 & 0.929 \\
& Flying with Photons & 0.243 & 0.695 & 1.227 & 0.262 & 0.715 & 1.185 & 0.599 & 0.465 & 1.270 \\
& Multilayer Imaging & \textbf{0.083} & \textbf{0.941} & \textbf{0.485} & \underline{0.154} & 0.822 & \underline{0.956} & \underline{0.270} & 0.532 & \underline{0.670} \\
& Ours & \underline{0.118} & \underline{0.883} & \underline{0.497} & \textbf{0.108} & \textbf{0.911} & \textbf{0.474} & \textbf{0.144} & \textbf{0.851} & \textbf{0.442} \\
\bottomrule
\end{tabular}
\vspace{-0.5em}
\end{table*}

\subsection{Multi-View Occlusion LiDAR Dataset}

\textbf{Acquisition system.}
We acquire single-photon transient data using an Adaps ADS6311 solid-state single-photon LiDAR system~\citep{adaps_hawk}. The sensor follows a direct time-of-flight configuration with time-correlated single-photon counting. A 940 nm VCSEL transmitter emits short laser pulses, and the returned photons are detected by a SPAD array. After on-chip spatial binning, each captured view provides a single-photon transient tensor of size \(192\times256\times672\)
\((H\times W\times T)\), with a temporal bin width of 750 ps. We additionally mount a Livox Avia LiDAR~\citep{livox_avia} on the capture rig to provide
reference geometry.

\textbf{Dataset collection.}
To support evaluation of layered reconstruction through partially
transmissive occluders, we build a real paired multi-view
single-photon LiDAR dataset. To the best of our knowledge, this is the first dataset that provides real multi-view SPL transient histograms
with paired occluded and clean captures at fixed sensor poses. The
dataset contains 30 scenes, formed by 3 occluding materials and 10
hidden-object categories. The occluding materials include a mosquito
net, a PVC shower curtain, and a black concealment mesh. For each scene, the occluder is mounted on a support frame, and
the hidden object is placed behind it. This setup produces mixed
single-photon transients, where the foreground occluder and hidden
object may contribute to the same sensor ray. For each scene, we
capture 16--19 views along an approximately \(180^\circ\) trajectory.
We use a fixed split for all methods, with 10 uniformly selected views
for training and the remaining views for testing.

\textbf{Calibration.}
We calibrate the SPL intrinsics using a checkerboard and obtain poses by transforming the Livox poses from Fast-Lio2~\citep{xu2022fast} into the SPL coordinate
frame. The effective temporal responses
\(\boldsymbol{\kappa}^{\mathrm{fg}}\) and \(\boldsymbol{\kappa}^{\mathrm{hid}}\) are estimated from training-view data. We select rays with clearly separated foreground and
hidden echoes, crop local waveform windows around each echo, align
them by the echo peak, and average them separately. Details are provided in the supplementary
material.

\textbf{Evaluation metrics.}
We report both depth and point-cloud metrics over three regions:
foreground, hidden, and mask-on-hidden. Foreground Depth L1 measures
the error between the predicted foreground depth and the reference
foreground depth from the occluded capture. Hidden Depth L1 measures
the error between the predicted hidden depth and the clean
reference depth over the full clean scene. Mask-on-Hidden Depth L1
uses the same hidden prediction but evaluates only within the manually
annotated hidden-object mask, directly measuring the accuracy on
the occluded object. For point-cloud evaluation, we use the corresponding foreground,
hidden, and mask-on-hidden regions and report Chamfer Distance (CD)
and F1 Score. We report F1@2, where the matching tolerance corresponds
to two temporal bins of the device histogram, approximately
22.5 cm in range. Together, these metrics evaluate pixel-level depth
accuracy and 3D geometric consistency for foreground--hidden
separation.

\subsection{Experimental setup}

\textbf{Baselines.}
We compare our method with three groups of baselines. The first group
includes neural transient reconstruction methods, \textbf{TransientNeRF}
\citep{malik2023transient}, \textbf{Transientangelo}
\citep{luo2025transientangelo}, and \textbf{Flying with Photons}
\citep{malik2024flying}, which optimize scene representations from
time-resolved SPAD histograms. The second is the \textbf{multilayer}
single-photon \textbf{imaging} method~\citep{halimi2017restoration}, which estimates multiple depth layers
from transient histograms without multi-view neural field
optimization. The third is \textbf{DyNFL}~\citep{huang2023neural}, a
point-cloud-based neural LiDAR method trained from processed point
observations. Since these baselines do not directly produce our foreground and hidden outputs, we use a unified layer-extraction protocol across all baselines. For neural transient baselines, we render the predicted transient at each test ray and extract the two strongest local peaks; the nearer peak is assigned to the foreground layer and the farther peak to the hidden layer. For rays with one reliable peak, the same depth is retained in both foreground-view and hidden scene outputs. For DyNFL, the input point observations are generated from the captured SPAD histograms using the same peak-extraction procedure, and its predicted near and far points are evaluated as foreground and hidden outputs. For the multilayer imaging baseline, we apply the method to training-view histograms, back-project the predicted layers from all
training views into a global point cloud, and project the accumulated
point cloud into each test view; the projected near and far layers are used as foreground and hidden predictions.

\begin{figure*}[!htbp]
    \centering
    \includegraphics[width=\columnwidth]{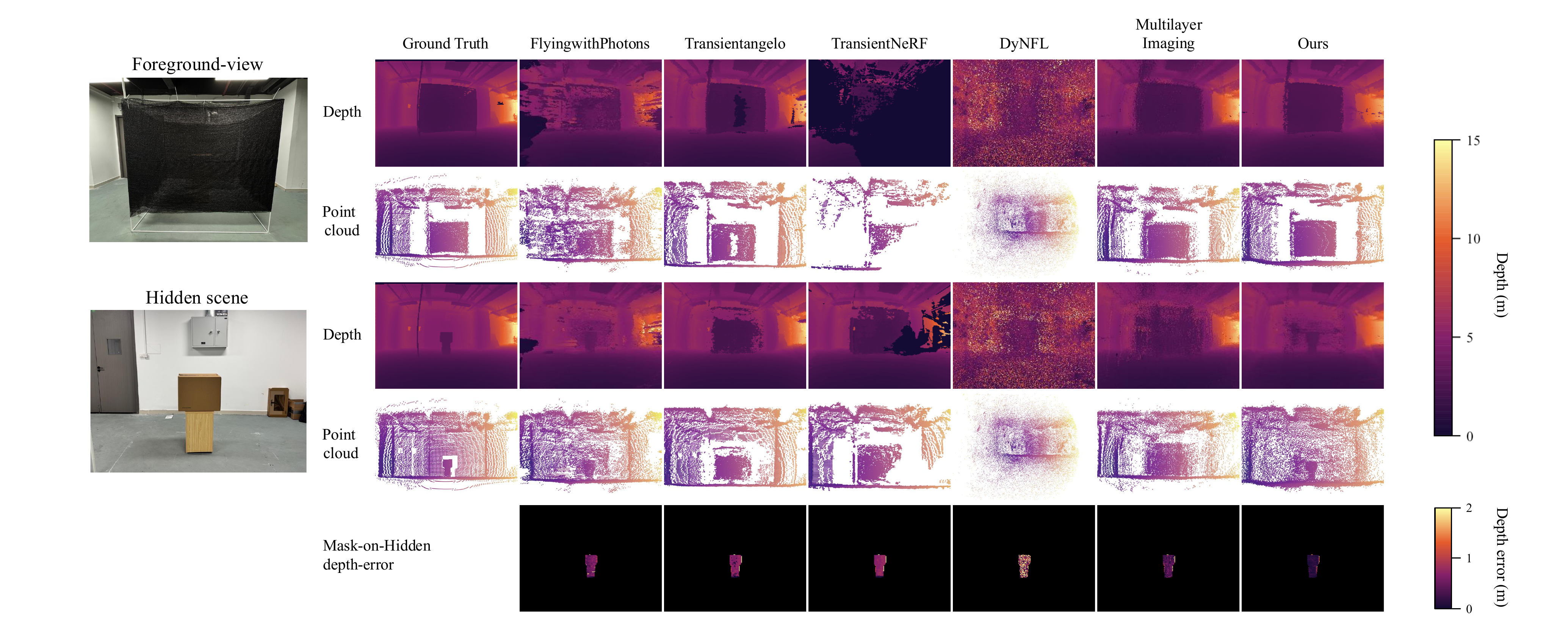}
    \caption{Qualitative comparison of foreground-view and hidden scene reconstructions across methods.}
    \label{fig:baseline_depth_pointcloud}
\end{figure*}

\subsection{Results}

The results also reveal the limitation of transient fitting as a
geometric objective. Neural transient baselines use full-waveform
supervision, but they optimize a single scene representation. When
foreground and hidden echoes are close, attenuated, or partially
overlapping, the measured waveform can be matched by a merged,
thickened, or wrongly assigned surface. Thus, a low transient-fitting
error does not necessarily imply correct foreground--hidden geometry.
DyNFL further illustrates the opposite limitation: once the histogram
is reduced to extracted range points, weak hidden echoes and waveform
shape cues are discarded, leading to noisy or incomplete hidden
predictions. Multilayer Imaging explicitly estimates multiple layers
and therefore performs well on strong foreground returns, but its
training-view point aggregation lacks a consistent multi-view layered
representation, limiting hidden-object accuracy in test views. The foreground metrics show the trade-off of layered reconstruction.
Multilayer Imaging achieves slightly better average foreground results,
which is expected because the foreground return is usually stronger
and easier to extract. Our method remains competitive on foreground
quality and achieves the best foreground results under the mosquito
net. More importantly, it substantially improves hidden and
Mask-on-Hidden reconstruction for all occluders. This trend shows that
good foreground recovery alone is not sufficient for the task: methods
that favor the dominant near return can reconstruct the occluder, but
still fail to recover the hidden object. Our method provides a better
foreground--hidden trade-off by explicitly using echo-state cues to
assign geometry to the two layers.

\textbf{Quantitative comparison.}
Table~\ref{tab:main_results} reports quantitative results on the
captured multi-view occlusion LiDAR dataset. Across the three occluders, our method consistently achieves the best hidden scene reconstruction. Compared with the strongest baseline on average, our method reduces Hidden CD from \(0.154\) to \(0.108\), improves Hidden F1@2 from \(0.822\) to \(0.911\), and reduces Hidden L1 from \(0.956\) m to \(0.474\) m. The gain is more pronounced on Mask-on-Hidden, where F1@2 improves from \(0.532\) to \(0.851\) and L1 decreases from \(0.670\) m to \(0.442\) m. This shows that the improvement comes from more accurate recovery of the occluded object itself, rather than only from better global hidden scene depth.

\textbf{Qualitative comparison.}
Fig.~\ref{fig:baseline_depth_pointcloud} supports the quantitative
findings. Transient baselines often produce incomplete or spatially
mis-assigned hidden structures, while DyNFL suffers from noisy point
clouds due to peak-based point extraction. Multilayer Imaging preserves
the foreground well, but its hidden reconstruction is often
incomplete or mis-localized. In contrast, our method recovers cleaner
hidden depth maps and more coherent hidden point clouds, with lower
hidden-mask errors indicating improvements on the truly occluded object
regions. These results reinforce our central claim: fitting the total transient does not guarantee correct layered geometry. A single representation may match the waveform while thickening or mis-assigning
foreground and hidden surfaces. By extracting per-ray local echo
evidence and using the inferred state to route waveform, localization,
and partition supervision, our method achieves more reliable
foreground--hidden reconstruction than transient fitting or
point-based baselines.

\begin{table}[!htbp]
\centering
\small
\setlength{\tabcolsep}{1.7pt}
\renewcommand{\arraystretch}{0.92}
\caption{Ablation study. Best results are in bold.}
\label{tab:ablation}
\begin{tabular}{@{}ccc@{\hspace{0.55em}}cc@{\hspace{0.55em}}cc@{\hspace{0.55em}}cc@{}}
\toprule
\multicolumn{3}{c}{Components}
& \multicolumn{2}{c}{Foreground}
& \multicolumn{2}{c}{Hidden}
& \multicolumn{2}{c}{Mask-on-Hidden} \\
\cmidrule(lr){1-3}
\cmidrule(lr){4-5}
\cmidrule(lr){6-7}
\cmidrule(l){8-9}
State & $\mathcal{L}_{\mathrm{loc}}$ & $\mathcal{L}_{\mathrm{sep}}$
& F1@2 $\uparrow$ & L1 (m) $\downarrow$
& F1@2 $\uparrow$ & L1 (m) $\downarrow$
& F1@2 $\uparrow$ & L1 (m) $\downarrow$ \\
\midrule
\checkmark &  & 
& 0.873 & 0.430
& 0.568 & 2.258
& 0.291 & 1.373 \\
 & \checkmark & \checkmark
& 0.852 & 0.557
& \textbf{0.926} & \textbf{0.427}
& \textbf{0.919} & 0.419 \\
\checkmark & \checkmark & 
& 0.884 & \textbf{0.402}
& 0.904 & 0.493
& 0.864 & 0.428 \\
\checkmark & \checkmark & \checkmark
& \textbf{0.901} & 0.419
& 0.916 & 0.459
& 0.863 & \textbf{0.415} \\
\bottomrule
\end{tabular}
\vspace{-0.6em}
\end{table}

\textbf{Ablation study.}
We conduct ablations on 10 randomly selected scenes, optimizing all variants for 30,000 iterations. As shown in Table~\ref{tab:ablation}, state-aware routing alone can fit the foreground well, but fails to recover the hidden scene: the hidden F1 drops to \(0.568\) and the hidden L1 error increases to \(2.258\) m. This suggests that routing the waveform supervision is not sufficient to disentangle layered geometry. Adding only the echo-evidence-based losses, \(\mathcal L_{\mathrm{loc}}\) and \(\mathcal L_{\mathrm{sep}}\), greatly improves hidden reconstruction, achieving the best hidden F1/L1 and Mask-on-Hidden F1. However, this
variant sacrifices foreground quality, reducing foreground F1 to \(0.852\) and increasing foreground L1 to \(0.557\) m. Combining state-aware routing with \(\mathcal L_{\mathrm{loc}}\) gives a more
balanced reconstruction and achieves the best foreground L1 (\(0.402\) m). The full method further adds
\(\mathcal L_{\mathrm{sep}}\), slightly increasing foreground L1 but
improving foreground F1, hidden F1, hidden L1, and Mask-on-Hidden L1
over the state+\(\mathcal L_{\mathrm{loc}}\) variant. These results
show that state-aware routing stabilizes foreground reconstruction,
while echo-evidence-based localization and separation are necessary
for recovering hidden geometry; their combination gives the best
overall trade-off.

\begin{figure}[!htbp]
    \centering
    \includegraphics[width=0.6\columnwidth]{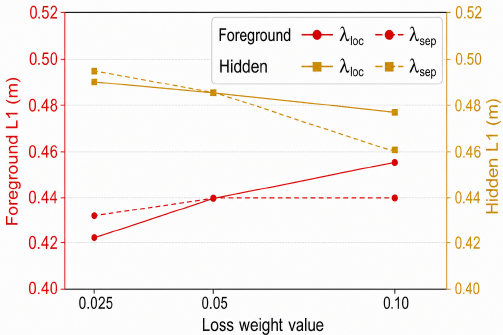}
    \caption{Sensitivity to loss weights. Foreground and hidden Depth L1 errors are
reported when varying \(\lambda_{\mathrm{loc}}\) and \(\lambda_{\mathrm{sep}}\). }
    \label{fig:sensitivity}
\end{figure}

\textbf{Hyperparameter sensitivity.}
We further evaluate the sensitivity of the localization and separation loss weights. As shown in Fig.~\ref{fig:sensitivity}, the method remains stable over the tested range. Increasing
\(\lambda_{\mathrm{loc}}\) slightly increases the foreground Depth L1,  but consistently reduces the hidden Depth L1, indicating that stronger anchor supervision helps recover the hidden scene while mildly affecting the foreground surface. Varying
\(\lambda_{\mathrm{sep}}\) shows a similar trend for the hidden layer, with only minor changes to the foreground error. Overall, the
performance varies smoothly with the loss weights, and the default
setting \(\lambda_{\mathrm{loc}}=\lambda_{\mathrm{sep}}=0.05\)
provides a balanced trade-off between foreground and hidden
reconstruction.

\section{Conclusion \& Limitations }

We introduced a state-aware single-photon framework for recovering
foreground-view and hidden scene from mixed transient
histograms. Our results show that matching the measured transient
alone is insufficient for layered reconstruction, as foreground and
hidden returns can be merged or mis-assigned while still fitting the
waveform. To resolve this, our method estimates local echo
evidence and routes waveform, localization, and partition supervision
to a two-head neural field. We also built a real paired SPL occlusion dataset with occluded/clean captures, calibrated poses, and hidden-object masks. Experiments show improved hidden scene and mask-on-hidden reconstruction over transient, multilayer, and point-based LiDAR baselines. The main limitation is that layer separation requires recoverable temporal evidence. If the hidden return is fully suppressed or indistinguishable from the foreground response, the layered geometry remains ambiguous. Future work may extend the framework to dynamic scenes and stronger scattering media.

\bibliography{main}
\bibliographystyle{tmlr}

\appendix

\section{Additional Results}

\subsection{Detailed Main Results}


We additionally apply the Multilayer Imaging algorithm~\citep{halimi2017restoration} to the histograms rendered by each transient-based baseline. Despite using a dedicated multilayer recovery method, the resulting foreground–hidden reconstructions remain substantially less accurate than ours as shown in Table~\ref{tab:fifth_peak_metrics}. This indicates that the main limitation is not the specific output extraction procedure: a single scene representation may already merge, broaden, or misassign the foreground and hidden components during transient optimization, after which post-hoc multilayer decomposition cannot reliably recover the underlying geometry.

Table~\ref{tab:fifth_peak_metrics} further shows that the improvement is stable across occluders and object instances. Our method consistently achieves low Hidden CD on black mesh, mosquito net, and PVC curtain, with \(0.107\pm0.002\), \(0.099\pm0.003\), and \(0.119\pm0.011\), respectively. It also maintains high Hidden F1@2 scores of \(0.907\pm0.003\), \(0.926\pm0.005\), and \(0.900\pm0.021\). The small standard deviations indicate that the gains are not dominated by a few easy cases, but hold across different captured scenes.

\begin{table*}[!htbp]
\centering
\scriptsize
\setlength{\tabcolsep}{3.5pt}
\renewcommand{\arraystretch}{1.08}
\newcommand{\best}[1]{\textbf{#1}}
\newcommand{\second}[1]{\underline{#1}}
\resizebox{\textwidth}{!}{%
\begin{tabular}{llccccccccc}
\toprule
\multirow{2}{*}{Occluder} & \multirow{2}{*}{Method} & \multicolumn{3}{c}{Foreground} & \multicolumn{3}{c}{Hidden} & \multicolumn{3}{c}{Mask-on-Hidden} \\
\cmidrule(lr){3-5} \cmidrule(lr){6-8} \cmidrule(lr){9-11}
 & & CD (m)$\downarrow$ & F1@2$\uparrow$ & L1 (m)$\downarrow$ & CD (m)$\downarrow$ & F1@2$\uparrow$ & L1 (m)$\downarrow$ & CD (m)$\downarrow$ & F1@2$\uparrow$ & L1 (m)$\downarrow$ \\
\midrule
\multirow{9}{*}{Black mesh} & DyNFL & 0.373$\pm$0.258 & 0.692$\pm$0.152 & 1.656$\pm$0.869 & 0.408$\pm$0.236 & 0.642$\pm$0.130 & 1.904$\pm$0.762 & 0.375$\pm$0.179 & 0.499$\pm$0.079 & 1.197$\pm$0.594 \\
 & Transientangelo & 0.194$\pm$0.057 & 0.841$\pm$0.039 & 1.768$\pm$0.238 & 0.155$\pm$0.025 & \second{0.857$\pm$0.013} & 1.244$\pm$0.179 & 0.495$\pm$0.187 & 0.525$\pm$0.082 & 0.840$\pm$0.259 \\
 & TransientNeRF & 0.299$\pm$0.120 & 0.748$\pm$0.085 & 0.986$\pm$0.334 & 0.179$\pm$0.050 & 0.838$\pm$0.034 & 1.092$\pm$0.242 & 0.415$\pm$0.198 & \second{0.546$\pm$0.112} & 0.967$\pm$0.306 \\
 & Flying with Photons & 0.252$\pm$0.102 & 0.634$\pm$0.086 & 1.064$\pm$0.259 & 0.241$\pm$0.061 & 0.705$\pm$0.053 & 1.056$\pm$0.184 & 0.578$\pm$0.183 & 0.453$\pm$0.075 & 1.386$\pm$0.122 \\
 & Multilayer Imaging & \best{0.076$\pm$0.001} & \best{0.952$\pm$0.001} & \best{0.427$\pm$0.007} & \second{0.134$\pm$0.006} & 0.846$\pm$0.008 & \second{0.873$\pm$0.029} & \second{0.245$\pm$0.037} & 0.491$\pm$0.105 & \second{0.642$\pm$0.096} \\
 & Transientangelo+MLI & 0.235$\pm$0.064 & 0.821$\pm$0.042 & 1.022$\pm$0.239 & 0.155$\pm$0.026 & \second{0.857$\pm$0.014} & 1.089$\pm$0.179 & 0.493$\pm$0.187 & 0.527$\pm$0.082 & 0.963$\pm$0.259 \\
 & TransientNeRF+MLI & 0.381$\pm$0.137 & 0.694$\pm$0.079 & 1.812$\pm$0.337 & 0.178$\pm$0.050 & 0.838$\pm$0.034 & 1.240$\pm$0.242 & 0.413$\pm$0.198 & \second{0.546$\pm$0.112} & 0.836$\pm$0.306 \\
 & FwP+MLI & 0.354$\pm$0.114 & 0.617$\pm$0.092 & 1.446$\pm$0.264 & 0.253$\pm$0.061 & 0.706$\pm$0.053 & 1.280$\pm$0.184 & 0.458$\pm$0.183 & 0.453$\pm$0.075 & 0.957$\pm$0.122 \\
 & Ours & \second{0.134$\pm$0.003} & \second{0.848$\pm$0.006} & \second{0.490$\pm$0.012} & \best{0.107$\pm$0.002} & \best{0.907$\pm$0.003} & \best{0.496$\pm$0.015} & \best{0.118$\pm$0.015} & \best{0.886$\pm$0.023} & \best{0.342$\pm$0.079} \\
\midrule
\multirow{9}{*}{Mosquito net} & DyNFL & 0.137$\pm$0.022 & 0.879$\pm$0.033 & 0.734$\pm$0.130 & 0.191$\pm$0.017 & 0.811$\pm$0.030 & 1.105$\pm$0.107 & \second{0.240$\pm$0.019} & 0.584$\pm$0.036 & 0.794$\pm$0.116 \\
 & Transientangelo & 0.199$\pm$0.118 & 0.846$\pm$0.087 & 1.883$\pm$0.458 & 0.160$\pm$0.056 & 0.861$\pm$0.039 & 1.261$\pm$0.260 & 0.550$\pm$0.191 & 0.399$\pm$0.139 & 0.956$\pm$0.235 \\
 & TransientNeRF & 0.309$\pm$0.180 & 0.724$\pm$0.134 & 1.100$\pm$0.628 & 0.186$\pm$0.039 & 0.836$\pm$0.035 & 1.004$\pm$0.312 & 0.512$\pm$0.171 & 0.434$\pm$0.085 & 0.987$\pm$0.225 \\
 & Flying with Photons & 0.276$\pm$0.048 & 0.688$\pm$0.055 & 1.417$\pm$0.166 & 0.304$\pm$0.038 & 0.691$\pm$0.041 & 1.281$\pm$0.135 & 0.665$\pm$0.092 & 0.413$\pm$0.102 & 0.957$\pm$0.353 \\
 & Multilayer Imaging & \second{0.078$\pm$0.008} & \second{0.951$\pm$0.012} & \second{0.428$\pm$0.041} & 0.170$\pm$0.012 & 0.805$\pm$0.009 & 0.995$\pm$0.033 & 0.243$\pm$0.027 & \second{0.709$\pm$0.027} & \second{0.617$\pm$0.102} \\
 & Transientangelo+MLI & 0.228$\pm$0.132 & 0.816$\pm$0.107 & 1.094$\pm$0.461 & \second{0.158$\pm$0.056} & \second{0.863$\pm$0.039} & \second{0.988$\pm$0.260} & 0.540$\pm$0.191 & 0.402$\pm$0.139 & 0.978$\pm$0.235 \\
 & TransientNeRF+MLI & 0.372$\pm$0.200 & 0.664$\pm$0.142 & 1.896$\pm$0.636 & 0.184$\pm$0.039 & 0.837$\pm$0.035 & 1.244$\pm$0.312 & 0.509$\pm$0.171 & 0.435$\pm$0.085 & 0.955$\pm$0.225 \\
 & FwP+MLI & 0.261$\pm$0.052 & 0.679$\pm$0.057 & 1.214$\pm$0.168 & 0.252$\pm$0.038 & 0.691$\pm$0.041 & 1.211$\pm$0.135 & 0.420$\pm$0.092 & 0.411$\pm$0.102 & 1.460$\pm$0.353 \\
 & Ours & \best{0.076$\pm$0.004} & \best{0.954$\pm$0.007} & \best{0.361$\pm$0.048} & \best{0.099$\pm$0.003} & \best{0.926$\pm$0.005} & \best{0.425$\pm$0.024} & \best{0.139$\pm$0.052} & \best{0.867$\pm$0.074} & \best{0.414$\pm$0.193} \\
\midrule
\multirow{9}{*}{PVC curtain} & DyNFL & \second{0.135$\pm$0.015} & \second{0.877$\pm$0.019} & 0.820$\pm$0.106 & 0.184$\pm$0.019 & 0.813$\pm$0.024 & 1.096$\pm$0.117 & \second{0.211$\pm$0.024} & 0.629$\pm$0.042 & \second{0.653$\pm$0.086} \\
 & Transientangelo & 0.292$\pm$0.128 & 0.765$\pm$0.086 & 2.126$\pm$0.352 & 0.171$\pm$0.070 & \second{0.859$\pm$0.047} & 1.325$\pm$0.260 & 0.530$\pm$0.245 & \second{0.781$\pm$0.098} & 0.919$\pm$0.304 \\
 & TransientNeRF & 0.407$\pm$0.151 & 0.693$\pm$0.109 & 1.368$\pm$0.409 & 0.192$\pm$0.034 & 0.854$\pm$0.024 & 1.095$\pm$0.218 & 0.573$\pm$0.221 & 0.685$\pm$0.069 & 0.834$\pm$0.225 \\
 & Flying with Photons & 0.202$\pm$0.014 & 0.765$\pm$0.018 & 1.201$\pm$0.052 & 0.241$\pm$0.018 & 0.748$\pm$0.020 & 1.216$\pm$0.054 & 0.555$\pm$0.083 & 0.531$\pm$0.081 & 1.466$\pm$0.182 \\
 & Multilayer Imaging & \best{0.095$\pm$0.001} & \best{0.920$\pm$0.001} & \best{0.601$\pm$0.005} & \second{0.157$\pm$0.003} & 0.815$\pm$0.003 & \second{0.999$\pm$0.009} & 0.321$\pm$0.045 & 0.396$\pm$0.052 & 0.750$\pm$0.102 \\
 & Transientangelo+MLI & 0.353$\pm$0.138 & 0.722$\pm$0.097 & 1.460$\pm$0.357 & 0.171$\pm$0.070 & \second{0.859$\pm$0.047} & 1.095$\pm$0.260 & 0.530$\pm$0.245 & \second{0.781$\pm$0.098} & 0.834$\pm$0.304 \\
 & TransientNeRF+MLI & 0.494$\pm$0.175 & 0.627$\pm$0.112 & 2.218$\pm$0.418 & 0.192$\pm$0.034 & 0.854$\pm$0.024 & 1.325$\pm$0.218 & 0.573$\pm$0.221 & 0.685$\pm$0.069 & 0.919$\pm$0.225 \\
 & FwP+MLI & 0.193$\pm$0.012 & 0.744$\pm$0.015 & 1.101$\pm$0.053 & 0.200$\pm$0.018 & 0.748$\pm$0.020 & 1.056$\pm$0.054 & 0.354$\pm$0.083 & 0.531$\pm$0.081 & 1.386$\pm$0.182 \\
 & Ours & 0.144$\pm$0.003 & 0.848$\pm$0.009 & \second{0.639$\pm$0.017} & \best{0.119$\pm$0.011} & \best{0.900$\pm$0.021} & \best{0.499$\pm$0.026} & \best{0.176$\pm$0.026} & \best{0.800$\pm$0.039} & \best{0.569$\pm$0.069} \\
\midrule
\multirow{9}{*}{Mean} & DyNFL & 0.217$\pm$0.139 & 0.815$\pm$0.108 & 1.075$\pm$0.517 & 0.262$\pm$0.129 & 0.755$\pm$0.100 & 1.371$\pm$0.472 & 0.277$\pm$0.089 & \second{0.570$\pm$0.067} & 0.885$\pm$0.288 \\
 & Transientangelo & 0.226$\pm$0.058 & 0.819$\pm$0.048 & 1.141$\pm$0.202 & 0.161$\pm$0.008 & \second{0.860$\pm$0.003} & 1.057$\pm$0.060 & 0.521$\pm$0.025 & \second{0.570$\pm$0.193} & 0.925$\pm$0.079 \\
 & TransientNeRF & 0.335$\pm$0.063 & 0.723$\pm$0.028 & 1.919$\pm$0.186 & 0.185$\pm$0.007 & 0.843$\pm$0.009 & 1.270$\pm$0.048 & 0.498$\pm$0.081 & 0.555$\pm$0.125 & 0.904$\pm$0.061 \\
 & Flying with Photons & 0.251$\pm$0.072 & 0.696$\pm$0.065 & 1.224$\pm$0.179 & 0.235$\pm$0.030 & 0.715$\pm$0.030 & 1.182$\pm$0.115 & 0.411$\pm$0.053 & 0.465$\pm$0.061 & 1.268$\pm$0.272 \\
 & Multilayer Imaging & \best{0.083$\pm$0.010} & \best{0.941$\pm$0.018} & \best{0.485$\pm$0.100} & \second{0.154$\pm$0.018} & 0.822$\pm$0.022 & \second{0.956$\pm$0.072} & \second{0.270$\pm$0.044} & 0.532$\pm$0.160 & \second{0.670$\pm$0.071} \\
 & Transientangelo+MLI & 0.272$\pm$0.070 & 0.786$\pm$0.055 & 1.192$\pm$0.235 & 0.161$\pm$0.008 & \second{0.860$\pm$0.003} & 1.057$\pm$0.060 & 0.521$\pm$0.025 & \second{0.570$\pm$0.193} & 0.925$\pm$0.079 \\
 & TransientNeRF+MLI & 0.416$\pm$0.068 & 0.661$\pm$0.034 & 1.975$\pm$0.214 & 0.185$\pm$0.007 & 0.843$\pm$0.009 & 1.270$\pm$0.048 & 0.498$\pm$0.081 & 0.555$\pm$0.125 & 0.904$\pm$0.061 \\
 & FwP+MLI & 0.269$\pm$0.081 & 0.680$\pm$0.064 & 1.254$\pm$0.176 & 0.235$\pm$0.030 & 0.715$\pm$0.030 & 1.182$\pm$0.115 & 0.411$\pm$0.053 & 0.465$\pm$0.061 & 1.268$\pm$0.272 \\
 & Ours & \second{0.118$\pm$0.037} & \second{0.883$\pm$0.061} & \second{0.497$\pm$0.138} & \best{0.108$\pm$0.010} & \best{0.911$\pm$0.013} & \best{0.474$\pm$0.041} & \best{0.144$\pm$0.030} & \best{0.851$\pm$0.045} & \best{0.442$\pm$0.116} \\
\bottomrule
\end{tabular}%
}
\caption{
Expanded quantitative results with additional baselines and standard
deviations. Results are grouped by occluder type and reported as
mean$\pm$std over scenes in each group. The best and second-best
results within each occluder block are bolded and underlined,
respectively. MLI denotes Multilayer Imaging, and FwP denotes Flying
with Photons.
}
\label{tab:fifth_peak_metrics}
\end{table*}

\subsection{Ablation of First-Photon Mechanism}

We ablate the first-photon mechanism by removing the windowed first-photon mapping in transient rendering. In this variant, the rendered arrival flux is used directly for waveform supervision, while the two-head field, echo-state routing, and anchor supervision remain unchanged. Table~\ref{tab:ablation_first_photon} compares the results with and without the first-photon mechanism. Removing the first-photon model leads to stronger foreground metrics: Foreground F1@2 increases from \(0.901\) to \(0.9449\), and Foreground L1 decreases from \(0.419\,\mathrm{m}\) to \(0.3685\,\mathrm{m}\). This is expected because the foreground return is usually strong and can be fitted well even with a simplified transient model. However, this improvement does not transfer to the hidden layer. With the first-photon mechanism, Hidden F1@2 improves from \(0.8962\) to \(0.917\), while Hidden L1 and CD decrease from \(0.4683\,\mathrm{m}\) to \(0.459\,\mathrm{m}\) and from \(0.1165\,\mathrm{m}\) to \(0.102\,\mathrm{m}\), respectively. For Mask-on-Hidden, the two variants give similar F1@2 and CD, while the first-photon model reduces L1 from \(0.4495\,\mathrm{m}\) to \(0.421\,\mathrm{m}\). These results show that the first-photon mechanism mainly benefits hidden-scene reconstruction rather than foreground fitting. Without this mechanism, the model can fit the dominant foreground return more directly, but it does not account for the suppression of delayed photons caused by earlier detections within the dead-time window. The first-photon model makes the waveform supervision more consistent with the actual SPL measurement process, which helps recover weaker hidden returns even though it slightly reduces foreground accuracy.

\begin{table*}[!htbp]
\centering
\small
\setlength{\tabcolsep}{4pt}
\resizebox{\textwidth}{!}{
\begin{tabular}{cccccccccc}
\toprule
\multicolumn{1}{c}{First Photon}
& \multicolumn{3}{c}{Foreground}
& \multicolumn{3}{c}{Hidden}
& \multicolumn{3}{c}{Mask-on-Hidden} \\
\cmidrule(lr){2-4}
\cmidrule(lr){5-7}
\cmidrule(lr){8-10}
& F1@2 $\uparrow$ & L1 (m) $\downarrow$ & CD (m) $\downarrow$
& F1@2 $\uparrow$ & L1 (m) $\downarrow$ & CD (m) $\downarrow$
& F1@2 $\uparrow$ & L1 (m) $\downarrow$ & CD (m) $\downarrow$ \\
\midrule
w/o
 & $0.9449 \pm 0.0101$ & $0.3685 \pm 0.0450$ & $0.0761 \pm 0.0137$
 & $0.8962 \pm 0.0346$ & $0.4683 \pm 0.0750$ & $0.1165 \pm 0.0195$
 & $0.8629 \pm 0.0600$ & $0.4495 \pm 0.2960$ & $0.1407 \pm 0.0506$ \\
w
 & $0.901 \pm 0.057$ & $0.419 \pm 0.080$ & $0.104 \pm 0.032$
 & $0.917 \pm 0.010$ & $0.459 \pm 0.040$ & $0.102 \pm 0.005$
 & $0.861 \pm 0.069$ & $0.421 \pm 0.193$ & $0.142 \pm 0.050$ \\
\bottomrule
\end{tabular}
}
\caption{Ablation Study of the First Photon Model.}
\label{tab:ablation_first_photon}
\end{table*}

\subsection{Local Echo Evidence Extraction}
\label{supp:echo_evidence_algorithm}

This section describes how we extract the per-ray echo state and
temporal anchors used by the state-aware supervision. The procedure
uses only the occluded training histograms and the calibrated response
templates.

\paragraph{Template models.}
For each ray \(r\), let \(\mathbf H_r=\{H_{r,b}\}_{b=1}^{B}\) be the
measured histogram and \(C_r=\sum_b H_{r,b}\) be the total recorded
count. We compare three explanations:
\[
\begin{aligned}
\boldsymbol{\Phi}_r^{\mathcal M_0}
&= \alpha_{\mathrm{amb}}\mathbf u,\\
\boldsymbol{\Phi}_r^{\mathcal M_1}
&= \alpha_s\mathbf R^{\mathrm{fg}}(a_s)
 + \alpha_{\mathrm{amb}}\mathbf u,\\
\boldsymbol{\Phi}_r^{\mathcal M_2}
&= \alpha_f\mathbf R^{\mathrm{fg}}(a_f)
 + \alpha_h\mathbf R^{\mathrm{hid}}(a_h)
 + \alpha_{\mathrm{amb}}\mathbf u .
\end{aligned}
\]
Here, \(\mathbf u\) is a unit-sum uniform temporal component for
ambient photons and dark counts, and
\(\mathbf R^{\mathrm{fg}}(a)\) and \(\mathbf R^{\mathrm{hid}}(a)\)
are shifted foreground/visible and hidden response templates centered
at temporal bin \(a\). All amplitudes are constrained to be
non-negative. For a fitted model \(\mathcal M_m\), let
\(\mathbf P_r^m=\mathcal S_{\mathrm{win}}
(\boldsymbol{\Phi}_r^{\mathcal M_m})\) be the resulting
detection-time distribution. We define the count-normalized multinomial negative log-likelihood as
\[
\ell_m(r)
=
-\sum_{b=1}^{B}
\frac{H_{r,b}}{C_r}
\log(P_{r,b}^{m}+\epsilon),
\]
where \(C_r=\sum_b H_{r,b}\) is the total recorded count of the
measured histogram, as defined in the main text. Thus,
\(C_r\ell_m(r)\) is the multinomial negative log-likelihood up to
constants independent of the model.

We then select the echo model using a BIC score
\[
\mathrm{BIC}_m(r)
=
2C_r\ell_m(r)
+
d_m\log(C_r),
\]
where \(d_m\) is the number of free fitted parameters in
\(\mathcal M_m\). We use \(d_0=1\), \(d_1=3\), and \(d_2=5\),
corresponding to: one ambient amplitude for \(\mathcal M_0\); one
return amplitude, one ambient amplitude, and one temporal center for
\(\mathcal M_1\); and two return amplitudes, one ambient amplitude,
and two ordered temporal centers for \(\mathcal M_2\). The model with
the lowest BIC score determines the echo state: \(\mathcal M_0\)
indicates no reliable surface evidence, \(\mathcal M_1\) provides a
single-return anchor \(a_r^s\), and \(\mathcal M_2\) provides ordered
anchors \((a_r^{\mathrm{fg}},a_r^{\mathrm{hid}})\).

\paragraph{Fast candidate search and fitting.}
The echo centers are used as histogram-bin anchors in the subsequent
geometry supervision. We therefore estimate them with a bin-level
candidate search followed by likelihood fitting. For each ray, we first
propose a compact set of plausible temporal centers from the measured
histogram. For each fixed center or ordered center pair, we optimize
the non-negative amplitudes, evaluate the windowed first-photon
likelihood, and finally select the echo state using the BIC score. This
keeps the fitting consistent with the discrete histogram representation
while avoiding exhaustive evaluation of all \(B\) single centers and
all \(O(B^2)\) ordered center pairs.

Given a histogram \(\mathbf H_r\), we smooth it along the temporal
axis, estimate a constant ambient-and-dark level, subtract it, and
clamp negative values to zero. This gives a nonnegative signal profile
\(\bar{\mathbf H}_r\). We then build the candidate set
\(\mathcal A_r\) from three sources. First, we include local-peak candidates. A bin is retained if it is a local maximum after non-maximum suppression and has sufficient
prominence and local photon mass. The prominence is measured relative
to the neighboring valleys within a local temporal window. The local
photon mass is
\(M_r(b)=\sum_{|j-b|\le w_m}\bar H_{r,j}\), which measures the signal
support around bin \(b\). Second, we include curvature/shoulder candidates. Weak hidden echoes
may appear as a shoulder or a slope change rather than as an isolated
peak. We therefore compute finite-difference derivative responses on
the smoothed profile and retain bins with strong local curvature or
slope-change response, subject to the same local-mass and
non-maximum-suppression checks. Third, we include support-guided candidates to cover broad or plateau-like responses that may not produce sharp local maxima. We identify the effective signal support as the connected temporal region where the smoothed signal or sliding-window mass remains above the ambient-and-dark level. Within this support, we add representative bins near the support start, midpoint, and end, the maximum-response bin, several offset bins around this maximum, and a small set of sparsely sampled bins across the support. These candidates ensure that the search covers the temporal extent of broad echoes, rather than only isolated peak or shoulder locations.

The union of these candidates forms \(\mathcal A_r\). Duplicate or
near-duplicate bins are merged, and the remaining candidates are ranked
by a score combining prominence, curvature response, and local photon
mass. We keep the top candidates for efficiency. The single-return
model \(\mathcal M_1\) is evaluated for \(a_s\in\mathcal A_r\), and
the two-return model \(\mathcal M_2\) is evaluated for ordered pairs
\((a_f,a_h)\in\mathcal A_r^2\) with \(a_f<a_h\) and a minimum
separation. For each candidate model instance, we fit the non-negative amplitudes
by minimizing the conditional multinomial negative log-likelihood after
the windowed first-photon mapping. The best instance of each model is
the one with the lowest likelihood among its candidates. We then
compute \(\mathrm{BIC}_m(r)=2C_r\ell_m(r)+d_m\log C_r\), where
\(C_r=\sum_b H_{r,b}\), \(\ell_m(r)\) is the count-normalized
multinomial negative log-likelihood, and \(d_m\) is the number of free
parameters. The model with the lowest BIC score gives the echo state
and anchors. If no valid candidate is found, the ray is assigned to
\(\mathcal M_0\), corresponding to no reliable surface evidence.

\subsection{Echo-State and Anchor Accuracy Validation}
\label{sec:echo-evidence-validation}

We validate the extracted local echo evidence by comparing its predicted echo states and temporal anchors with annotations on occluded transient histograms. The evaluation covers all three occluder types in our dataset: black mesh, mosquito net, and PVC shower curtain. For each occluder type, we randomly sample and annotate 100 spatial points across multiple views. Each annotation specifies whether the occluded histogram contains single-return or two-return evidence, together with the corresponding temporal anchor locations. These annotations are used only for evaluation and are not provided to the echo fitting procedure. Given only the occluded histogram, the method fits competing single-return and two-return explanations under the windowed first-photon model, and predicts the echo state and temporal anchors. A prediction is evaluated against the annotated state; anchor errors are reported in temporal bins for the corresponding annotated returns. Table~\ref{tab:echo_state_anchor_accuracy} reports echo-state classification accuracy and anchor localization errors. The method achieves \(96.0\%\) overall state accuracy and \(97.6\%\) two-return recall, showing that the occluded waveform usually provides sufficient evidence to distinguish single-return and two-return cases. Anchor localization is most accurate for the black mesh, with foreground and hidden errors of \(0.63\) and \(0.93\) bins. For the mosquito net, the foreground anchor remains accurate (\(0.93\) bins), while the hidden anchor error increases to \(2.44\) bins. The shower curtain gives perfect state classification, but larger foreground and hidden anchor errors of \(4.24\) and \(5.66\) bins. These results indicate that echo-state selection is robust across the three occluding materials, while precise anchor localization is more sensitive to the temporal response of the material. Overall, the evaluation supports using occluded histograms to provide reliable state supervision.

\begin{table}[!htbp]
\centering
\resizebox{\columnwidth}{!}{
\begin{tabular}{lccccc}
\toprule
Material &
State Acc. &
Single Recall &
Two Recall &
FG MAE &
Hidden MAE \\
\midrule
Blacknet & 93.9\% & 80.0\% & 96.4\% & 0.63 & 0.93 \\
Mosquito net & 93.9\% & 80.0\% & 96.4\% & 0.93 & 2.44 \\
Shower & 100.0\% & 100.0\% & 100.0\% & 4.24 & 5.66 \\
\midrule
Overall & 96.0\% & 86.7\% & 97.6\% & 1.99 & 3.07 \\
\bottomrule
\end{tabular}
}
\caption{Accuracy of echo-state classification and foreground/background anchor localization across occluding materials. Anchor localization errors are reported in temporal bins.}
\label{tab:echo_state_anchor_accuracy}
\end{table}

\subsection{Cross-View Anchor Projection Accuracy}

We further evaluate whether the extracted temporal anchors provide geometrically meaningful cross-view cues. This experiment directly projects the anchors into 3D, without optimizing a neural field. For each training-view ray with valid echo evidence, we convert the temporal anchor to range and project it along the corresponding LiDAR ray. Foreground anchors are accumulated into a foreground-view point cloud, and hidden anchors are accumulated into a hidden-scene point cloud. Shared single-return anchors are included in both outputs, following the definition of our foreground-view and hidden-scene reconstructions. The accumulated point clouds are then projected into test views and evaluated with the same foreground, hidden, and mask-on-hidden metrics. Table~\ref{tab:ablation_anchor_projection} compares direct anchor projection with our full reconstruction model. Directly projecting the extracted anchors already provides non-trivial foreground and hidden geometry, achieving Hidden F1@2 of \(0.689\) and Mask-on-Hidden F1@2 of \(0.353\). This shows that the estimated anchors contain meaningful cross-view geometric cues. However, direct anchor projection is still far from the full reconstruction quality. Compared with projected anchors, our full model improves Foreground F1@2 from \(0.765\) to \(0.901\), Hidden F1@2 from \(0.689\) to \(0.917\), and Mask-on-Hidden F1@2 from \(0.353\) to \(0.861\). The corresponding L1 errors are also substantially reduced, from \(1.010\) m to \(0.419\) m for foreground, from \(1.075\) m to \(0.459\) m for hidden regions, and from \(0.947\) m to \(0.421\) m on Mask-on-Hidden. This gap highlights the role of the neural reconstruction stage. The anchors are sparse, view-dependent, and may contain errors caused by response mismatch, weak hidden returns, or imperfect state selection. Using them directly as points does not enforce waveform consistency, continuous multi-view geometry, or foreground--hidden partitioning. In contrast, our full method uses the anchors as supervision for a two-head neural field, while still optimizing the transient likelihood across all views. Therefore, the experiment supports our design choice: local echo anchors provide useful geometric supervision, but they should guide a multi-view state-aware neural reconstruction rather than be used as the final reconstruction.

\begin{table*}[!htbp]
\centering
\small
\setlength{\tabcolsep}{4pt}
\resizebox{\textwidth}{!}{
\begin{tabular}{cccccccccc}
\toprule
\multicolumn{1}{c}{}
& \multicolumn{3}{c}{Foreground}
& \multicolumn{3}{c}{Hidden}
& \multicolumn{3}{c}{Mask-on-Hidden} \\
\cmidrule(lr){2-4}
\cmidrule(lr){5-7}
\cmidrule(lr){8-10}
& F1@2 $\uparrow$ & L1 (m) $\downarrow$ & CD (m) $\downarrow$
& F1@2 $\uparrow$ & L1 (m) $\downarrow$ & CD (m) $\downarrow$
& F1@2 $\uparrow$ & L1 (m) $\downarrow$ & CD (m) $\downarrow$ \\
\midrule
Anchor Projection
 & $0.765 \pm 0.020$ & $1.010 \pm 0.100$ & $0.195 \pm 0.020$
 & $0.689 \pm 0.012$ & $1.075 \pm 0.053$ & $0.238 \pm 0.010$
 & $0.353 \pm 0.067$ & $0.947 \pm 0.110$ & $0.442 \pm 0.053$ \\
Ours
 & $0.901 \pm 0.057$ & $0.419 \pm 0.080$ & $0.104 \pm 0.032$
 & $0.917 \pm 0.010$ & $0.459 \pm 0.040$ & $0.102 \pm 0.005$
 & $0.861 \pm 0.069$ & $0.421 \pm 0.193$ & $0.142 \pm 0.050$ \\
\bottomrule
\end{tabular}
}
\caption{
Cross-view projection accuracy of extracted temporal anchors. We
directly back-project training-view anchors into foreground-view and
hidden-scene point clouds, project them to test views, and evaluate
with the reconstruction metrics.
}
\label{tab:ablation_anchor_projection}
\end{table*}

\subsection{Automatic Estimation of Temporal Response Templates}
\label{supp:response_estimation}

Our local echo fitting and transient rendering use two effective
temporal response templates,
\(\boldsymbol{\kappa}^{\mathrm{fg}}\) and
\(\boldsymbol{\kappa}^{\mathrm{hid}}\), for foreground/visible and
hidden returns. These templates are estimated once per scene from the
occluded training measurements only. No paired clean capture,
hidden-object mask, or manual peak annotation is used.

For each scene, we first select a single training view
\(v^\star\) for response estimation. This view is chosen automatically
as the training view with the largest number of valid two-echo
candidate pixels under the screening procedure below. All remaining
steps are performed only on this selected view, and the resulting
templates are fixed for all views of the scene.

Given the histogram \(\mathbf H_r\) of each ray \(r\) in
\(v^\star\), we smooth the histogram along the temporal axis and
estimate a constant ambient-and-dark level from low-count bins. After
subtracting this level and clamping negative values to zero, we detect
local maxima with non-maximum suppression. A ray is selected as a
two-echo candidate if it contains two temporally ordered peaks
\(b_r^{\mathrm{fg}} < b_r^{\mathrm{hid}}\) that satisfy three
conditions: the two peaks are separated by at least a minimum temporal
gap, both peaks have sufficient prominence above the estimated
ambient-and-dark level, and both peaks contain sufficient local photon
mass within a small temporal window. These criteria remove flat
background rays, single-return rays, and noisy weak maxima.

For every selected two-echo candidate, we crop two local waveform
windows centered at \(b_r^{\mathrm{fg}}\) and
\(b_r^{\mathrm{hid}}\). The early crop is used as a foreground/visible
response sample, and the late crop is used as a hidden response sample.
Each crop is aligned to a common temporal center using its local peak
or centroid, normalized to unit area, and then added to the
corresponding response pool. We further reject outlier crops whose
width or correlation to the provisional mean is outside the accepted
range. The final templates are obtained by averaging the remaining
aligned and normalized crops:
\begin{equation}
\begin{aligned}
\boldsymbol{\kappa}^{\mathrm{fg}}
&= 
\operatorname{norm}
\left(
\frac{1}{|\mathcal R|}
\sum_{r\in\mathcal R}
\mathbf g_r^{\mathrm{fg}}
\right),\\
\boldsymbol{\kappa}^{\mathrm{hid}}
&= 
\operatorname{norm}
\left(
\frac{1}{|\mathcal R|}
\sum_{r\in\mathcal R}
\mathbf g_r^{\mathrm{hid}}
\right),
\end{aligned}
\label{eq:kappa}
\end{equation}
where \(\mathcal R\) is the set of selected two-echo candidate rays,
\(\mathbf g_r^{\mathrm{fg}}\) and \(\mathbf g_r^{\mathrm{hid}}\) are
the aligned local crops, and \(\operatorname{norm}(\cdot)\) denotes
normalization to unit sum. The estimated responses capture the
effective temporal broadening in the current scene and are used in
both template-based echo evidence estimation and neural transient
rendering.

\begin{figure}[!htbp]
    \centering
    \includegraphics[width=0.95\columnwidth]{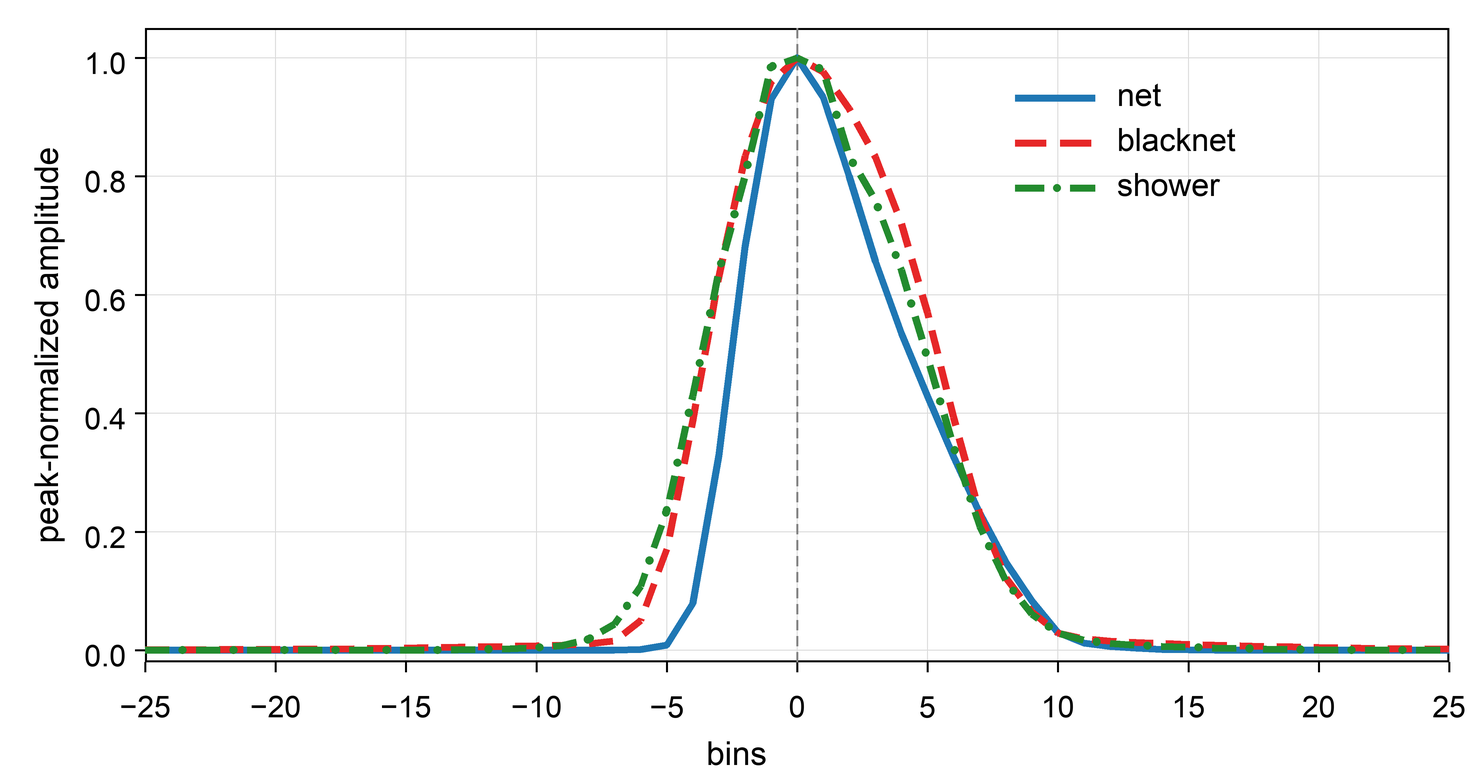}
\caption{ Automatically estimated foreground/visible response templates \(\boldsymbol{\kappa}^{\mathrm{fg}}\) from first echoes of different occluding materials. The profiles are temporally aligned and peak-normalized for shape comparison. }
    \label{fig:first_peak}
\end{figure}

\begin{figure}[!htbp]
    \centering
    \includegraphics[width=0.95\columnwidth]{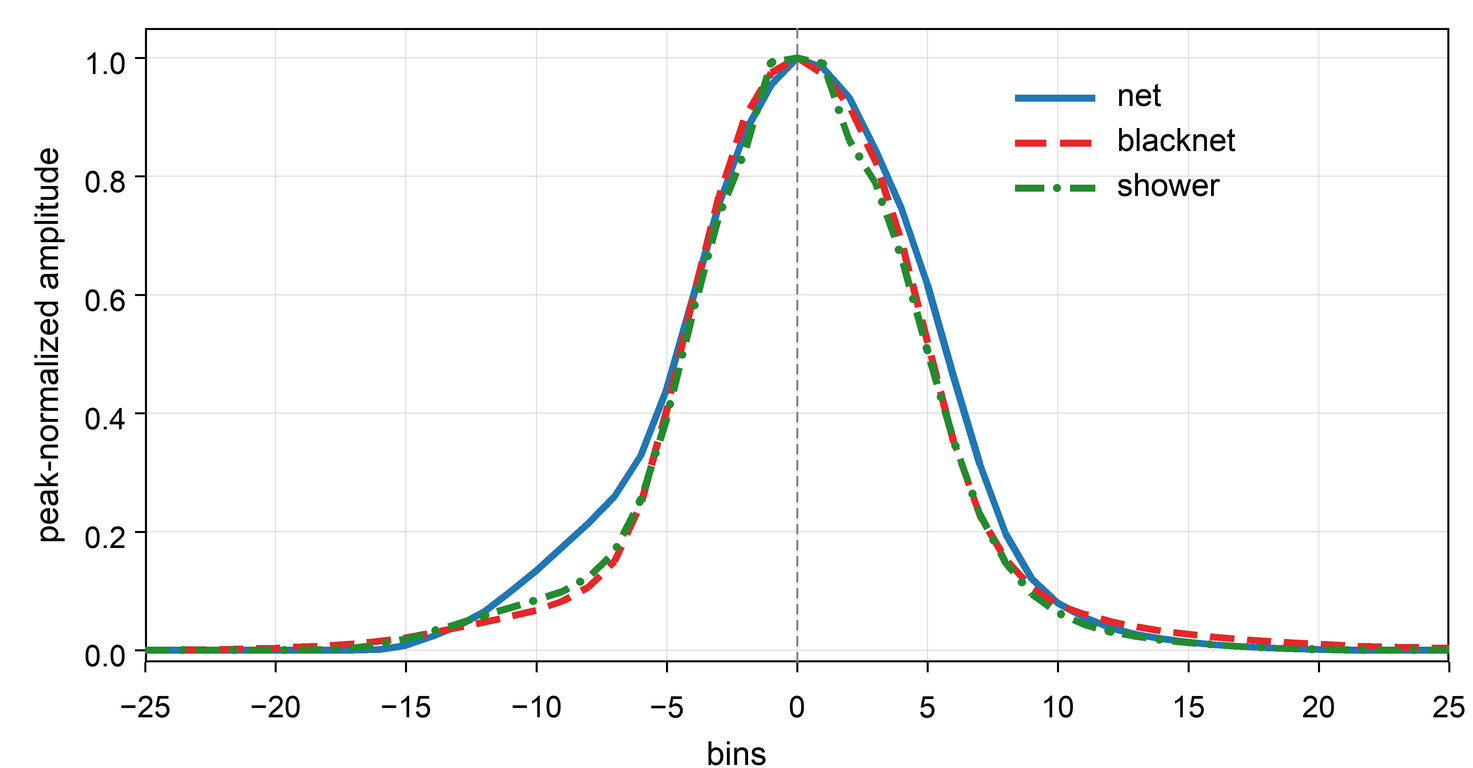}
\caption{ Automatically estimated hidden-response templates \(\boldsymbol{\kappa}^{\mathrm{hid}}\) from second echoes of different occluding materials. Compared with the first-echo templates, the hidden responses are broader and have longer temporal tails. }
    \label{fig:second_peak}
\end{figure}

\textbf{Response-template Results.}
Figures~\ref{fig:first_peak} and~\ref{fig:second_peak} show the automatically estimated, peak-normalized response templates for the three occluding materials. The first-echo profiles correspond to the foreground/visible response \(\boldsymbol{\kappa}^{\mathrm{fg}}\), while the second-echo profiles correspond to the hidden response \(\boldsymbol{\kappa}^{\mathrm{hid}}\). Since all profiles are temporally aligned and peak-normalized, the plots compare response shape rather than absolute return strength or echo delay. The estimated profiles show a consistent difference between the two returns. The foreground/visible responses are relatively narrow, with sharper rising and falling edges. In contrast, the hidden responses are broader and exhibit longer temporal tails, indicating additional temporal broadening after light passes through the occluder and returns from the hidden surface. The variation across occluding materials is visible but moderate compared with the foreground--hidden difference. This supports our use of separate effective response templates for foreground/visible and hidden returns, while estimating them automatically for each scene to capture material-dependent response changes.

\section{Implementation details}
\label{sec:impl}

Following the TransientNeRF setting~\citep{malik2023transient}, our backbone uses a multi-resolution hash-grid encoder with 16 levels, 2 features per level, base resolution 16, and maximum resolution 4096. We use an occupancy grid of resolution \(128^3\) for accelerated ray sampling and render up to 4096 samples per ray. The encoded position is passed to a shared one-hidden-layer MLP of width 64, which outputs a 15-dimensional geometric feature. Each input transient tensor has size \(192\times256\times672\) \((H\times W\times T)\), with temporal bin width \(7.5\times10^{-10}\) s. Rays are rendered over the range \([0,16]\) m. Each training batch contains 16 rays, with a target sample budget of 65,536 samples. We train each scene for 100,000 steps using Adam with learning rate \(10^{-3}\) and random seed 42. Our method modifies this transient field with a state-aware layered formulation. On top of the shared representation, we use separate foreground and hidden prediction branches. Two independent one-hidden-layer density decoders predict \(\sigma^{\mathrm{fg}}\) and \(\sigma^{\mathrm{hid}}\), and two independent two-hidden-layer radiance decoders predict the corresponding return amplitudes \(\rho^{\mathrm{fg}}\) and \(\rho^{\mathrm{hid}}\), conditioned on the geometric feature and view direction. The rendered arrival profile is passed through the windowed first-photon sensor model with a dead-time window of 13.5 temporal bins~\citep{adaps_hawk}. For local echo-evidence extraction, the foreground-template detection probability is set to 0.2. The state-aware objective uses \(\lambda_{\mathrm{loc}}=0.05\) and \(\lambda_{\mathrm{sep}}=0.05\) for localization and partition supervision, respectively.

\section{Computation Time}

We report the average computation time per scene on the black mesh subset using 10 training views. All timings were measured on a workstation equipped with an NVIDIA RTX 4090 GPU and an Intel(R) Xeon(R) Silver 4214R CPU @ 2.40GHz with 48 threads. TransientNeRF, Transientangelo, Flying with Photons, Multilayer Imaging, DyNFL, and our method take 1.93, 2.77, 4.88, 0.26, 3.07, and 3.95 hours per scene, respectively. Multilayer Imaging is substantially faster because it performs per-view restoration and fusion rather than optimizing a neural scene representation; however, as shown in the main results, this efficiency comes with weaker hidden-scene consistency. Our method is slower than single-representation transient baselines such as TransientNeRF and Transientangelo because it additionally estimates echo states and optimizes foreground and hidden neural heads with localization and separation losses. Nevertheless, its runtime remains below Flying with Photons and within the same practical range as DyNFL, while providing substantially better hidden and mask-on-hidden reconstruction accuracy.

\section{Additional Qualitative Results}

\begin{figure*}[!htbp]
    \centering
    \includegraphics[width=\linewidth]{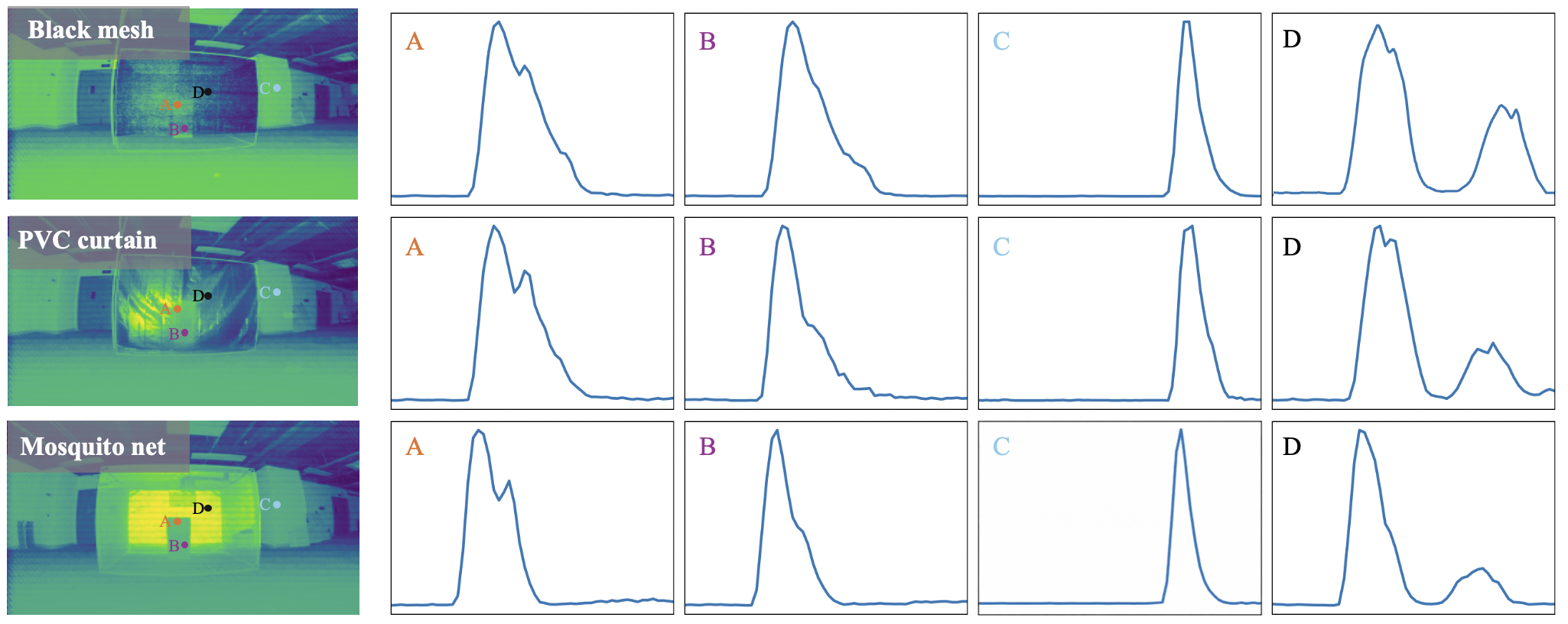}
    \caption{Material- and location-dependent transient waveforms under occlusion. For each occluding material, we mark four representative spatial locations and show their measured transient histograms. Different locations produce different echo structures, ranging from narrow single returns to broadened, asymmetric, or clearly separated multi-return waveforms. The waveform shape also changes with the occluding material. }
    \label{fig:waveforms_different_scenes}
\end{figure*}

\textbf{Visualization of Transient Waveform}
Fig.~\ref{fig:waveforms_different_scenes} visualizes measured transient histograms at representative spatial locations under different occluding materials. The waveforms vary substantially across both location and material. Some rays contain a relatively narrow single-return peak, while others show broadened peaks, asymmetric tails, shoulder-like structures, or two separated components. This variation reflects the heterogeneous visibility of the foreground occluder and the hidden scene along different rays. It also shows that a uniform peak-extraction rule is insufficient for layered reconstruction. Our echo-evidence extraction instead estimates the state and temporal anchors locally for each ray, allowing the supervision to adapt to the available waveform evidence.

\begin{figure*}[!htbp]
    \centering
    \includegraphics[width=\textwidth]{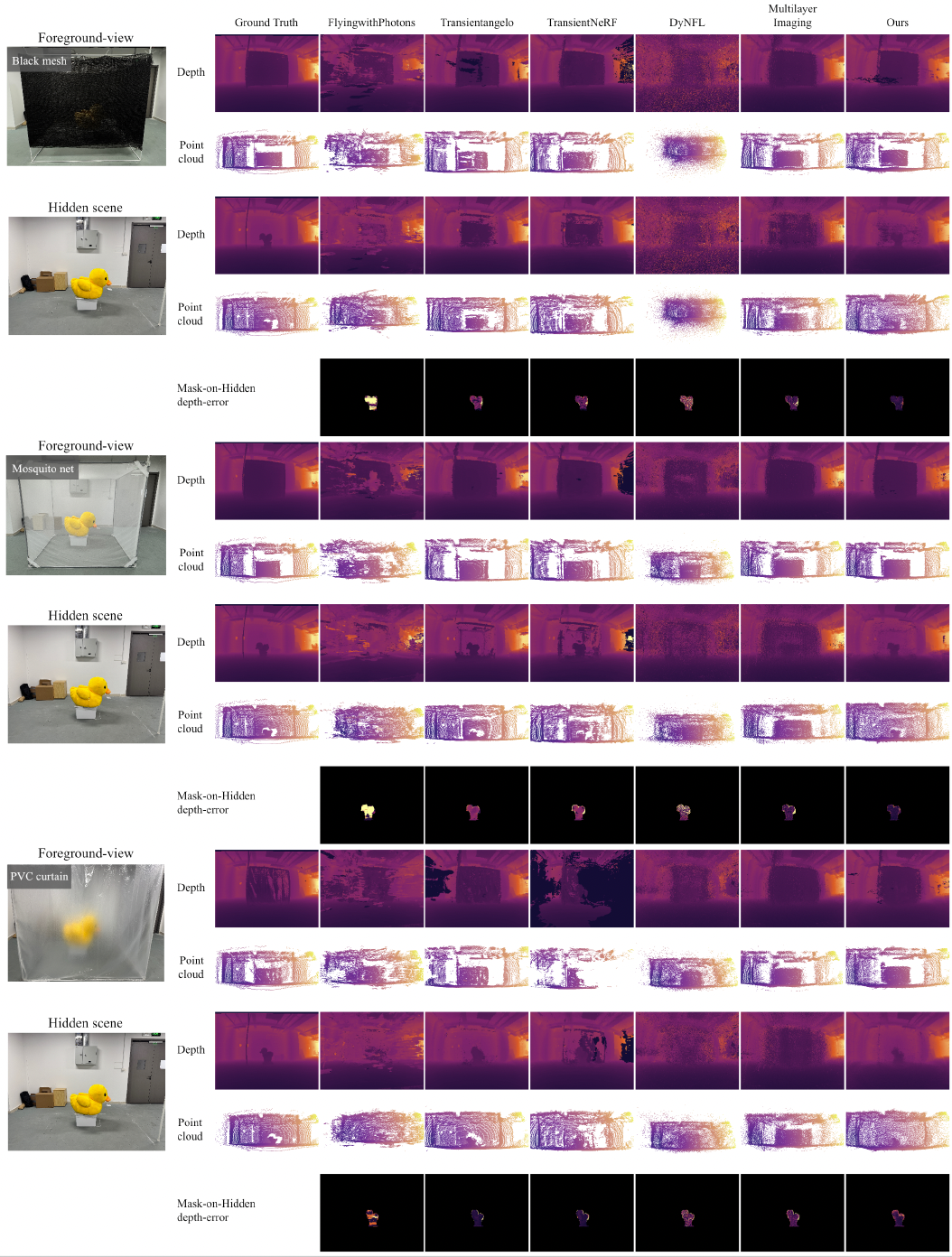}
    \caption{
    Additional qualitative comparison on the duck scene under different foreground occluders. 
    Depth maps and depth-colored point clouds are visualized within 0--15 m, while mask-on-hidden depth-error maps are visualized within 0--2 m.
    }
    \label{fig:supp_duck}
\end{figure*}

\begin{figure*}[!htbp]
    \centering
    \includegraphics[width=\textwidth]{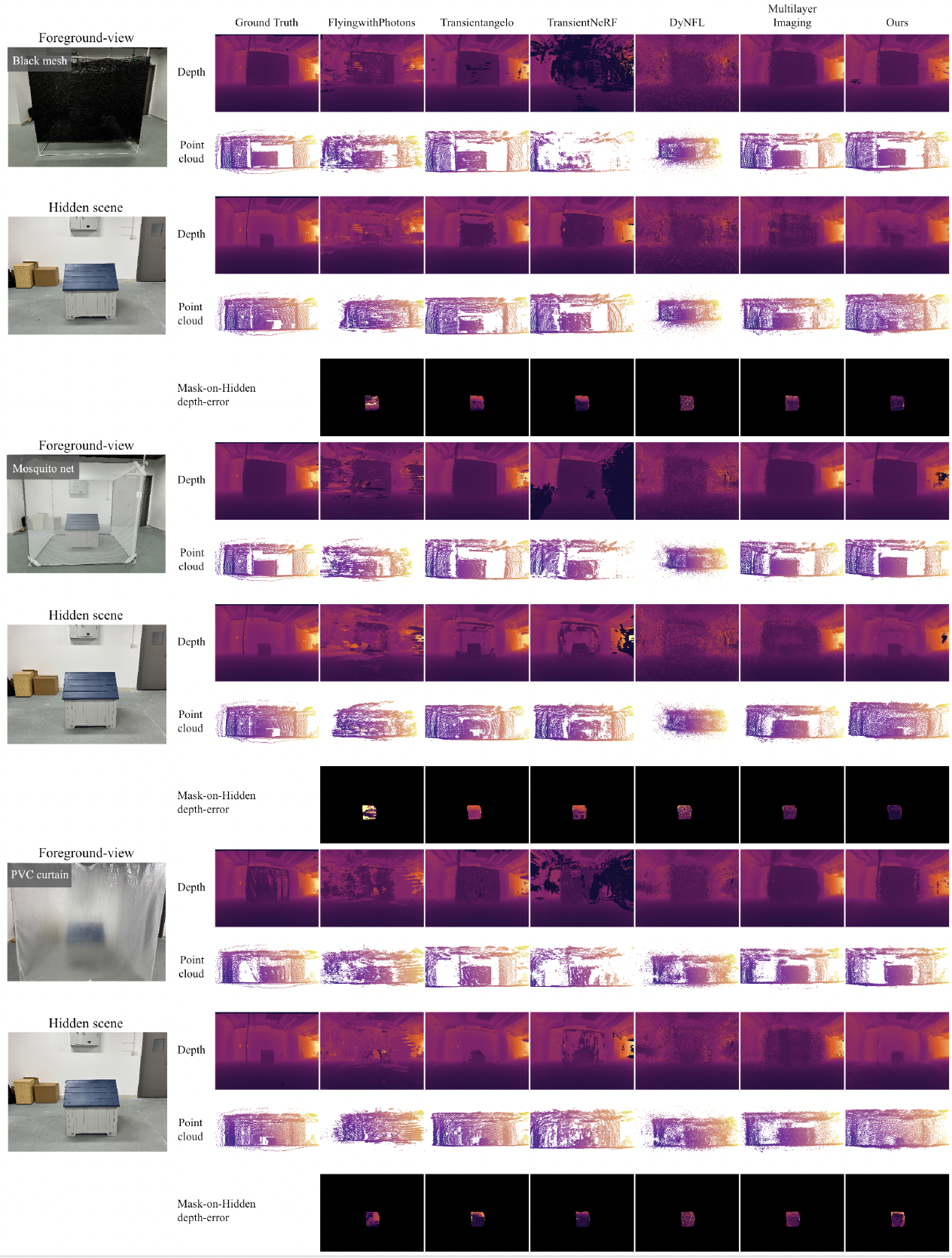}
    \caption{
    Additional qualitative comparison on the house scene under different foreground occluders. 
    Depth maps and depth-colored point clouds are visualized within 0--15 m, while mask-on-hidden depth-error maps are visualized within 0--2 m.
    }
    \label{fig:supp_house}
\end{figure*}

\begin{figure*}[!htbp]
    \centering
    \includegraphics[width=\textwidth]{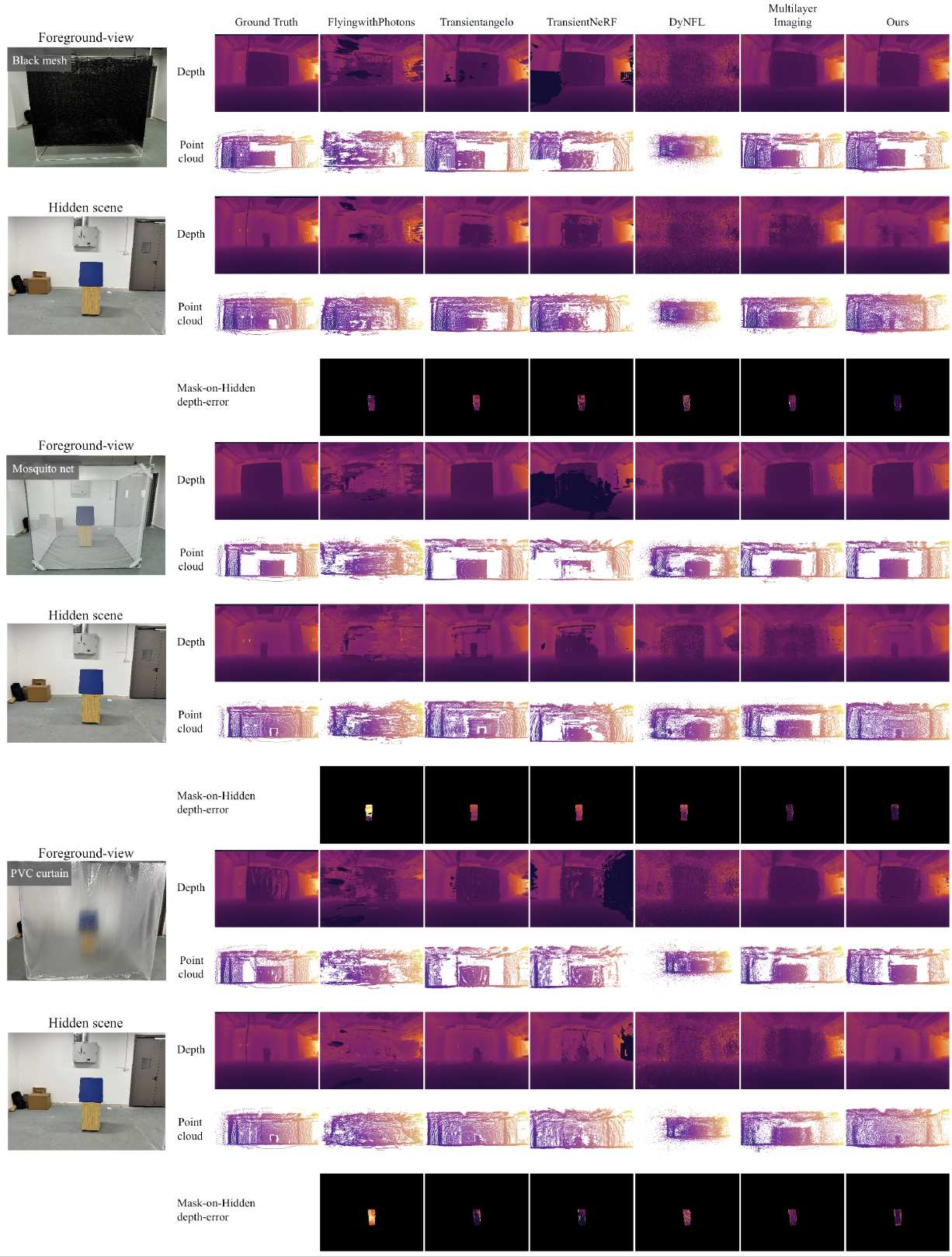}
    \caption{
    Additional qualitative comparison on the triangle scene under different foreground occluders. 
    Depth maps and depth-colored point clouds are visualized within 0--15 m, while mask-on-hidden depth-error maps are visualized within 0--2 m.
    }
    \label{fig:supp_triangle}
\end{figure*}

\textbf{Additional Qualitative Foreground--Hidden Reconstruction Results}
We provide additional qualitative comparisons for three hidden-scene categories: duck, house, and triangle as shown in Fig.~\ref{fig:supp_duck}-\ref{fig:supp_triangle}. For each category, we report results under three representative foreground occluders, including black mesh, mosquito net, and PVC curtain. Each comparison includes the foreground-view RGB image, the hidden-scene RGB image, the predicted depth maps, the corresponding depth-colored point clouds, and the mask-on-hidden depth-error maps. For all depth maps and depth-colored point clouds, the colormap is normalized to the same depth range of 0--15 m. For the mask-on-hidden depth-error maps, the colormap is normalized to 0--2 m. This separate normalization highlights fine-grained reconstruction errors in the hidden target region while preserving a consistent depth visualization range for the scene-level geometry.

\end{document}